\documentclass[lettersize,journal]{IEEEtran}
\usepackage[numbers,sort&compress]{natbib}

\usepackage{amsmath}
\usepackage{array,diagbox,siunitx}

\usepackage{pdflscape}
\usepackage{graphicx}
\usepackage{subcaption}
\usepackage{epstopdf}
\usepackage{subfig}
\usepackage{caption}

\usepackage{cases}
\usepackage{float}
\usepackage{amssymb}
\usepackage{multirow}
\usepackage{mathrsfs}
\usepackage{algorithmic}
\usepackage{rotating}
\usepackage{balance}
\usepackage{threeparttable}
\usepackage{natbib}
\usepackage{color}
\usepackage{colortbl}
\usepackage{makecell}

\definecolor{mygray}{gray}{.9}
\usepackage{makecell}
\usepackage{pdflscape}
\usepackage{siunitx}
\usepackage[pagewise]{lineno}
\usepackage{stfloats}
\usepackage{verbatim}
\usepackage{setspace}
\usepackage{threeparttable}
\usepackage{supertabular,booktabs}
\usepackage{rotating}
\usepackage{supertabular}
\usepackage{pifont}
\usepackage{xcolor,pict2e}% to allow any radius
\usepackage{algorithm}
\usepackage{algorithmic}
\usepackage{tikz}
\usepackage{booktabs}
\usepackage{footnote}
\usepackage[pagewise]{lineno}
\usepackage{balance}
\newcommand{\etal}{\textit{et al.}}

\newcommand{\eg}{\textit{e.g.}}

\begin{document}

\title{A Survey of Deep Learning for Group-level Emotion Recognition}

\author{Xiaohua Huang, %~\IEEEmembership{Senior Member,~IEEE},
	Jinke Xu,
	Wenming Zheng,~\IEEEmembership{Senior Member,~IEEE},
	Qirong Mao,%,
	Abhinav Dhall
	% <-this % stops a space
	
	\thanks{X. Huang is with the School of Computer Engineering, Nanjing Institute of Technology, China, the Key Laboratory of Child Development and Learning Science (Southeast University), Ministry of Education, Southeast University, Nanjing 210096, China, and also with Research Center for Learning Science, Southeast University, China. 
		(email: xiaohuahwang@gmail.com)}
	
	\thanks{J. Xu is with the School of Computer Engineering, Nanjing Institute of Technology, China. 
		(email: y00450220246@njit.edu.cn)}

	\thanks{W. Zheng is with the Key Laboratory of Child Development and Learning Science (Southeast University), Ministry of Education, Southeast University, Nanjing 210096, China and also with the School of Biological Science and Medical Engineering, Southeast University, Nanjing 210096, Jiangsu, China.
		(E-mail: wenming\_zheng@seu.edu.cn)}
	
	\thanks{Q. Mao is with the School of Computer Science and Communication Engineering, Jiangsu University, Zhenjiang, Jiangsu, China. (E-mail: mao\_qr@ujs.edu.cn)}
	
	\thanks{A. Dhall is with College of Science \& Engineering, Flinders University, Adelaide, Australia and Indian Institute of Technology Ropar, Hussainpur, Rupnagar 140001, Punjab. (E-mail: abhinav.dhall@flinders.edu.au)}
% <-this % stops a space
%\thanks{This work was supported in part by the National Natural Science Foundation of China under Grants 62076122 and 62176106, in part by the Talent Startup project of NJIT under Grant YKJ201982.}
%\thanks{Manuscript received April 19, 2021; revised August 16, 2021.}
}

\markboth{Journal of \LaTeX\ Class Files,~Vol.~14, No.~8, August~2021}%
{Shell \MakeLowercase{\textit{et al.}}: A Sample Article Using IEEEtran.cls for IEEE Journals}


\maketitle

\begin{abstract}
With the advancement of artificial intelligence (AI) technology, group-level emotion recognition (GER) has emerged as an important area in analyzing human behavior. Early GER methods are primarily relied on handcrafted features. However, with the proliferation of Deep Learning (DL) techniques and their remarkable success in diverse tasks, neural networks have garnered increasing interest in GER. Unlike individual's emotion, group emotions exhibit diversity and dynamics. Presently, several DL approaches have been proposed to effectively leverage the rich information inherent in group-level image and enhance GER performance significantly. In this survey, we present a comprehensive review of DL techniques applied to GER, proposing a new taxonomy for the field cover all aspects of GER based on DL. The survey overviews datasets, the deep GER pipeline, and performance comparisons of the state-of-the-art methods past decade.  Moreover, it summarizes and discuss the fundamental approaches and advanced developments for each aspect. Furthermore, we identify outstanding challenges and suggest potential avenues for the design of robust GER systems. To the best of our knowledge, thus survey represents the first comprehensive review of deep GER methods, serving as a pivotal references for future GER research endeavors.
\end{abstract}

%
%\section{Introduction}
%\IEEEPARstart{T}{his} file is intended to serve as a ``sample article file''

\section{Introduction}
% ~\textcolor{red}{According to~\cite{mehrabian1974approach}, facial expressions convey 55\% of the total amount of information in human daily communication, while voices and language convey 38\% and 7\% respectively}. 
% This emotion plane represents emotional states through continuous values on two dimensions, with the horizontal axis representing valence ranging from negative to positive, and the vertical axis representing arousal ranging from low to high, with the middle being neutral. This new way of expressing emotions provides a new perspective for better understanding and researching emotions. 

\IEEEPARstart{E}{motio} exhibits a profound influence on human perception, attention, memory, and decision-making,  directly impacting both physical and mental well-being~\cite{brosch2013impact}. Consequently, the comprehension and perception of emotions not only enhance interpersonal communication but also implicitly contribute to the regulation of human physical health. With the evolution of big data technology, the broadening of application scenarios, and the progress in artificial intelligence technology, group-level emotion recognition (GER) has become a focal point for researchers~\cite{dhall2017individual,wang2022congnn}. Unlike individual-level emotion recognition, which analyzes the emotions of individuals through facial expressions, speech, and postures, GER focuses on identifying the collective emotions of a group of people. Existing GER technology integrates diverse information, encompassing facial expressions, postures, social interactions, etc., to predict the emotions of a group. Taking facial expression data as an example, GER typically involves three key steps: first, given an image, detecting and extracting information such as faces and postures from images or videos; secondly, employing handcrafted descriptor or neural networks to extract facial features and other relevant information; finally, inputting these features as sequential data into another model like recurrent neural network to classify the group-level emotion. However, GER presents three key challenges. Firstly, the groups and contexts involved in GER may exhibit diversity and complexity. Secondly, the labeling of group-level emotion demands more nuanced considerations for the emotions expressed by the primary group, introducing additional complexities compared to individual-level emotion labeling. Lastly, due to the involvement of multiple individuals and diverse contexts, the recognition process becomes more complex than individual-level emotion recognition. Despite these challenges, GER remains an essential and formidable research area. It would be utilized to analyze emotion changes of a group of people, detecting abnormal behaviors and potential dangers in a timely manner in surveillance videos, or understanding students' learning status in collaborative learning~\cite{dindar2020leaders}.

%group-level emotion recognition infers the emotional tendency of a group by analyzing the emotional expressions and contextual information of group members in multimodal data. 

This article provides a comprehensive survey of deep learning approaches for GER. Different from existing review~\cite{veltmeijer2021automatic}, this paper elaborates on the methods of GER with a unique focus on deep learning architecture, providing insights into the current state and technical challenges associated with deep learning method in GER. In the beginning,we commence by providing an in-depth analysis of groups and emotions from a social perspective, offering a concise concept for GER. Subsequently, we present a thorough description of the image-based and video-based group-level emotion databases currently available for GER. Moreover, we explore recently developed deep learning methods for GER and scrutinize their technical challenges. In conclusion, we discuss the development trends of GER, offering valuable insights and guidance for future research and applications.

%the principles, methods, classifications, and applications of group-level emotion recognition based on deep networks, summarizes existing deep network methods, analyzes their technical challenges, and explores the 

\section{Group-level emotion}

Emotions, traditionally viewed as individual-level phenomena in common parlance, have increasingly received attention in the field of social psychology as being potentially group-level~\cite{smith2016group}. Niedenthal and Brauer offer a broad definition of group-level emotions, characterizing them as emotions experienced by individuals on behalf of a group with which they identify~\cite{niedenthal2012social}. This perspective suggests that group-level emotions are inherently more complex than their individual-level counterparts. Adopting both sociological and psychological perspectives, we will proceed by elucidating the concepts of groups and emotions before providing a precise definition of group emotions.

%in common sense have broadly been considered as a individual-level phenomenon. Now a range of research and theory in social psychology has introduced the concept of group-based emotion~\cite{smith2016group}. Niedenthal and Brauer broadly define group-based emotion as an emotion experienced by individuals on behalf of a group to which they belong and with which they identify, a definition that captures what is common among all these perspectives~\cite{niedenthal2012social}. Therefore, compared with individual-level phenomenon, group-level emotion become more complicated interpreted. According to a sociological and psychological perspective, we will start to describe the concept of group and emotion, and then offer the definition of group emotion.}

%According to a sociological and psychological perspective, this section offers a comprehensive description of group and emotion. %Subsequently, these two concepts are combined to explore the concept of group emotion in depth, in order for readers to have a deeper understanding of group emotion. This section  discusses the two constituents of group emotion: group and emotion. 

%\subsection{The concept of group}
The fundamental disparity between group-level emotion and individual-level emotion arises from the distinction between the concepts of a  group and an individual. While an individual pertains to an singular, independent entity, a group transcends mere aggregation, constituting a social phenomenon that amalgamates individuals into a collective entity. Various perspectives and definitions of the concept of a group have been proposed in scholarly literature~\cite{shaw1961group,szilagyi1983organizational,barron1994time,dasgupta1999group}. Shaw~\etal~\cite{shaw1961group} defines a group as comprising two or more individuals engaged in mutual interaction and influence, emphasizing the significance of interaction among group members. Szilagi and Wallance~\cite{szilagyi1983organizational} characterize a group as a collection of two or more individuals who interact and depend on each other to achieve a common goal. Additionally, Barron~\etal~\cite{barron1994time} further elaborate on this definition by conceptualizing a group as individuals connected by some kind of bond, exhibiting varying degrees of cohesion. Moreover, Dasgupta~\etal~\cite{dasgupta1999group} describe a group as a group of people closely interconnected in some manner. According to these studies, a group consists of multiple individuals with a shared objective and direct or indirect interactions between them during a certain period of time. Thus, only when these conditions are met can a combination of multiple individuals be deemed a group.

%\subsection{The concept of emotion}
Numerous scholars have investigated the concept of emotion within the sociological and psychological field. Schachter~\etal~\cite{schachter1962cognitive}  proposed that emotion encompasses both a physiological arousal state and a cognitive state  adapted to this physiological state. Kleinginna~\etal~\cite{kleinginna1981categorized} conducted a comprehensive review categorizing 92 different definitions of emotion from various literature sources. Ekman in~\cite{ekman1992argument} suggested that ``emotions evolve from the adaptive value of human beings in handling basic life tasks'', indicating that emotions arise as adaptive responses to different situations encountered during social activities. Averill regards emotion as a complex of impulsive motivation in higher cognition, involving various psychological processes and physiological responses~\cite{averill1998emotions}. Cabanac defines emotion as any psychological experience with high intensity and high pleasure content, emphasizing the relationship between emotion and psychological experience~\cite{cabanac2002emotion}. Barrett posits that emotions are constructs of the brain's interpretation of external and internal stimuli~\cite{barrett2006emotions}. Scherer views emotion as a biopsychological phenomenon resulting from the interaction of specific neural systems and physiological responses, cognitive assessments, and social-cultural factors~\cite{Schuller2010}. Therefore, emotions can be understood as physiological and psychological responses to stimuli in our environment.

%\subsection{Group emotion}
Based on the foregoing elucidation of group and emotion, group-level emotion denotes the physiological and psychological responses elicited by multiple interacting individuals over a defined period. Scholars have proffered diverse definitions of group-level emotion in their studies. In the beginning, Hatfield~\etal~\cite{hatfield1993emotional} characterize group-level emotion as the process whereby group members reciprocally influence each other through emotional contagion and empathy, culminating in emotional synchronization and consistency. Essentially, group-level emotion entails the reciprocal transmission and influence of emotions among individuals, leading to emotional consistency. Furthermore, group emotion is perceived as the emotional state propagated among members of a social group, cultivated and diffused through social interaction and shared experiences~\cite{bagozzi2006antecedents}.  Bars{\"a}de and Gibson stressed the importance for researchers in the social science community to approach group-level emotions from both a ``top-down approach'' and a ``bottom-up approach''~\cite{barsade1998group}. A ``top-down approach" suggests that emotion exhibited by group is represented at the group level and is felt by individual members, while the ''bottom-up approach'' highlights the unique compositional effects of individual-level group member emotions. According to the framework~\cite{barsade1998group}, Kelly and Bars{\"a}de further proposed that group-level emotion comprises affective compositional effects and affective context~\cite{kelly2001mood}. In essence, group-level emotion emerges from  a combination of individual-level affective factors posed by group members and group-level factors that shape the emotional experience of the group. Additionally, Bars{\"a}de and Gibson  further explore group-level emotion as the emotional state disseminated and shared among members of an organization~\cite{barsade2007does}.

\section{Group-level emotion dataset}

The rise of social media platforms has led to a surge in the volume of uploaded photos and videos, driving advancements in big data technologies and affective computing domains, especially GER. In recent years, numerous group-level emotion datasets are established, attracting attention from researchers in the domains of affective computing and computer vision. However, the quality of annotation on images and videos plays an critical role in determining the efficacy of GER models. Thus, this section aims to provide an exhaustive examination of existing group-level emotion datasets, classifying them into two main types: image-based and video-based types. %we will present a comprehensive discussion of group emotion datasets, categorizing them into image-based and video-based types. 

\subsection{Image-based datasets}

In contrast to individual-level emotion datasets, group-level emotion datasets necessitate annotation for a group. Image-based datasets began to emerge since 2013, with notable representations including  MultiEmoVA~\cite{mou2015group}, Happiness Image database (HAPPEI)~\cite{dhall2013finding}, Group-level Affect database (GAF)~\cite{dhall2017individual,dhall2015more,dhall2018emotiw}, Group cohesion dataset~\cite{ghosh2019predicting}, GroupEmoW~\cite{guo2020graph}, and SiteGroEmo~\cite{wang2022congnn}. Among these database, HAPPEI, GAF, and the Group cohesion dataset, which were utilized in the EmotiW sub-challenge, have received significant attention and adoption by researchers in the fields of computer vision and affective computing. 

Mou~\etal~\cite{mou2015group} introduced a multi-dimensional group emotion image dataset, namely MultiEmoVA.  This dataset was primarily constructed by scouring various social media platforms for real-life photos. Initially, 400 color images were collected, and after manually filtering images with ambiguous emotional expressions, 250 images meeting the criteria were remained. Furthermore, these images were categorized into positive, neutral, and negative. Additionally, Mou~\etal~also incorporated arousal-level annotation into the dataset, providing more explicit representations of the intensity for each emotion category.

The HAPPEI database contains images captured from social media platform and categorizes group-level emotions into neutral, small smile, big smile, smile, big laugh, and thrilled, totaling 2,638 images. In contrast, the GAF datbase delineates three emotion categories along the valence dimension: positive, neutral, and negative. The data collection process involved web search using keywords corresponding to various scenarios, such as weddings, birthday parties, and sports events, etc., to obtain images depicting group-level emotions in those contexts. Subsequently, these images were annotated by 2 to 3 experts in affective computing domain. The GAF database has undergone three iterations, namely GAF~\cite{dhall2015more}, GAF2.0~\cite{dhall2017individual}, and GAF3.0~\cite{dhall2018emotiw}. The initial version, GAF includes 504 images~\cite{dhall2015more}, which is relatively small in scale and has been limited usage by researcher for evaluating the performance of deep GER models. However, the subsequent iterations, GAF2.0 and GAF3.0, witnessed a substantial increase in data size, featuring 6,467~\cite{dhall2017individual} and 17,172 images~\cite{dhall2018emotiw}, respectively. %The data size of the last two versions has significantly increased, with 

In addition to emotion category, the cohesiveness of a group serves as a crucial indicator of the emotional state, structure and success of a group of individuals. Group Cohesion database was derived from GAF3.0 by adding cohesion labels~\cite{ghosh2019predicting}. Each image in this database was annotated with a cohesion score ranging from 0 to 3 by five annotators, where 0 represents no cohesion and 3 means strong cohesion. This database was utilized in the Emotiw2019 challenge~\cite{dhall2019emotiw}, with 9,300 images allocated for training, 4,244 images for validation and 2,899 images for testing purposes.

Besides the above-mentioned databases, another databases have recently proposed by Guo~\etal~\cite{guo2020graph} and Wang~\etal~\cite{wang2022congnn}, that is GroupEmoW and SiteGroEmo. GroupEmoW was created with a stringent criterion mandating that each image contains 2 to 9 individuals engaged in a specific activity, thereby forming distinct groups. According to this criterion, they collected 15,894 images from the internet, creating a diverse dataset with varying image resolutions and in-the-wild. Subsequently, these images were categorized into positive, neutral, and negative. On the other hand, the SiteGroEmo dataset diverges from  existing database by capturing image across tourism scenes worldwide. This dataset not only contains rich geographic information and scene variations but also randomly captures the facial and body movements of individuals at a specific moment. Comprising a total of 10,034 images, the dataset is labeled with valence to denote emotions, specifically categorized as negative, neutral, and positive.

It is important to note that the aforementioned image-based datasets predominantly rely on internet keywords searches for data collection. While this method may expedite the establishment of satisfactory datasets, the existing datasets, with the exception of the MultiEmoVA database, solely employ valence-level annotations for images, lacking arousal-level annotations and more specific emotion categories such as anger and surprise. This limitation may arise from the diverse expression exhibited by participants in a group, making it challenging to annotate group-level images with a comprehensive emotion taxonomy. Additionally, manual annotation may introduce discrepancies in labeling due to cultural differences.  Moreover, many images in these datasets may suffer from quality issues such as poor lighting, incomplete capture of facial expressions, or obstructions obscuring certain individuals. Finally, given their static nature, these images lack information about emotional dynamics. In the domain of emotion recognition research, scholars have emphasized that dynamic changes in facial expressions can offer crucial clues for both humans and computers to discern emotions or emotional processes~\cite{pantic2006dynamics}. Therefore, these uncontrollable circumstances and the absence of dynamic information may have a discernible impact on the accuracy of GER based on image.

\subsection{Video-based datasets}

In contrast to image-based datasets, video-based datasets not only capture the temporal dynamics of emotional expressions but also provide additional contextual information, facilitating a more comprehensive and accurate portrayal of changes in emotional states. However, due to the demanding collection and annotation process, only two video-based datasets have emerged recently since 2019: the VGAF dataset introduced by Sharma~\etal~\cite{sharma2019automatic} and the GECV dataset by Quach~\etal~\cite{quach2022non}. %Both datasets were collected from online platforms and categorized images into positive, neutral, or negative. 

The VAGF dataset comprises videos sourced from the YouTube platform, each featuring a varying number of individuals forming groups of different sizes. The dataset is partitioned into training, validation, and test sets, encompassing 2,661, 766, and 756 samples, respectively. Alongside valence annotation, the dataset includes cohesion metrics among individuals in the group. The corresponding dataset has also been used in EmotiW2023~\cite{dhall2023emotiw}. Conversely, the GECV dataset contains videos captured in leisure and crowded scenes, amounting to a total of 627 videos. This dataset is further subdivided into three subsets: GEVC-SingleImg, GEVC-GroupImg, and GEVC-GroupVid. The latter two subsets are tailored to capture group emotions more effectively by showcasing various various group behaviors across different scenarios.

While video-based databases offer richer semantic and contextual information compared to image-based databases, facilitating more discriminant criteria for data annotation, they are sourced from media platforms featuring complex scenes that often depict real-life scenarios. Nonetheless, they present limitations such as the absence of physiological signals akin to multi-modal emotion recognition, thereby posing challenges in data collection. %In the future, it becomes imperative to simulate relevant scenarios in laboratory environments to construct a comprehensive group emotion database encompassing facial, vocal, and physiological signals.

\subsection{Dataset summary}
The specific comparisons of the existing group emotion datasets are presented  in Table~\ref{tab:databaseintro}. Despite all datasets collecting images or videos from diverse real-world scenarios, their sample size is comparably small when compare with comprehensive datasets like AffectNet~\cite{mollahosseini2017affectnet}. This limited data size impedes the robust learning of group-level features. Recently, in the domain of micro-expression recognition, a composite dataset consolidating various micro-expression databases has become popular~\cite{see2019megc}. This approach corroborates the generalization capacity of the method across datasets with disparate characteristics, mitigating the issue of data scarcity. Such a strategy holds promise for GER by potentially augmenting data volumes, particularly for video-based datasets, and enhancing the generalization ability of GER methods.

State-of-the-art approaches, particularly those showcased in the EmotiW challenge, primarily undergo evaluation using  image-based databases. All databases adopt an annotation strategy categorizing images into three valence-level categories, as certain nuanced emotions such as fear and contempt pose challenges in data collection, resulting in limited samples for these categories, which are insufficient for robust learning. As seen from Table~\ref{tab:databaseintro}, except SiteGroEmo, each database has a balanced distribution of data across each class. Furthermore, in practical experiments, apart from the HAPPEI and MultiEmoVA databases, other databases follow an official protocol where the train, validation, and test sets remain strictly fixed without random validation. Such rigid dataset partitioning may not be facilitate accurate assessments of GER method performance.

%only the emotion categories with more than 10 samples are considered. Recently, the composite dataset is popular, because it can verify the generalization ability of the method on datasets with different natures. For further increasing the MER performance, MMEW collected micro- and macro-expressions from the same subjects which may be helpful for further cross-modal research.

% and example samples are shown in Figure 2. 

\begin{table*}
\centering
\caption{Group-level emotion database.}
\scriptsize
\label{tab:databaseintro}
\vspace{0.2cm}
\begin{tabular}{|c|c|c|c|c|c|}
	
	\hline
	Dataset & Type & Sample size & Category & Task &Protocol\\
	\hline
	HAPPEI~\cite{dhall2015more}* & Image & 2,638  & \makecell[c]{Neural (92), Small smile (147), \\Large smile (774), Small laugh (1256), \\Large laugh (331), Thrilled (38)} & Regression &\makecell[c]{4-fold cross validation \\ \textit{train} (1500), \textit{val} (1138), \textit{test} (496)}\\
	
	\hline
	MultiEmoVA~\cite{mou2015group} & Image & 250  & \makecell[c]{High-\textit{pos} (46), Medium-\textit{pos} (64), \\High-\textit{neg} (31), medium-\textit{neg} (27), \\low-\textit{neg} (10), \textit{neu} (72)} & Classification & 5-fold cross validation\\
	
	\hline
	%SocEID~\cite{ahsan2017complex} & Image & 37,000 images & arousal \& valence & \\

	%EmotiC~\cite{kosti2017emotion} & Image & 18,316 images & \makecell[c]{Happiness (), Surprise (), Aversion, Anger, Sadness, Fear\\ valence, arousal, and dominate} & \\
	GAF2.0~\cite{dhall2017individual} & Image & 6,467  & \textit{pos} (2,356), \textit{neu} (2,092), \textit{neg} (2,019) & Classification & \textit{train} (3,630), \textit{val} (2,068), \textit{test} (772) \\ % data information from Figure 4. "Emotion recognition in the wild using deep neural networks and bayesian classfiers, ICMI2017"
	
	\hline
	
	GAF3.0~\cite{dhall2018emotiw} & Image & 17,172  & \textit{pos} (6,553), \textit{neu} (5,364), \textit{neg} (5,256) & Classification & \textit{train} (9,836), \textit{val} (4346), \textit{test} (3011)\\
	
	\hline
	
	\textcolor{black}{Group Cohesion~\cite{dhall2019emotiw,ghosh2019predicting,ghosh2020automatic}} & Image & 16,433  & [0, 3] & Regression & \textit{train} (9,300), \textit{val} (4,244), \textit{test} (2,899)\\
	
	\hline
	
	SiteGroEmo~\cite{wang2022congnn} & Image & 10,034  & \textit{pos} (4,660), \textit{neu} (4,355), \textit{neg} (1,019) & Classification & \textit{train} (6,096), \textit{val} (1,972), \textit{test} (1,966)\\
	
	\hline
	
	GroupEmoW~\cite{guo2020graph} & Image & 15,894  & \textit{pos} (6,636), \textit{neu} (4,947), \textit{neg} (4,311) & Classification & \textit{train} (11,127), \textit{val} (3,178), \textit{test} (1,589)\\
	
	%\hline
	%GECV-GroupImg~\cite{quach2022non} & Image & 438000 & \textit{pos} (), \textit{neu} (), \textit{neg} () & \\
	
	% GroupEmoW [4] is a public GER dataset that consists of 15,894 images. It is divided into train, validation, and test sets, each with 11,127, 3,178, and 1,589 images. These images are collected from the Google, Baidu, Bing, and Flickr websites by searching for keywords related to social events, such as funeral, birthday, protest, conference, meeting, and wedding. The collective emotions of the images are also labeled with negative, neutral, or positive valence state. Fig. 7(b) shows some examples from the GroupEmoW dataset.
	
	%\hline
	
	%HECO~\cite{yang2022emotion} & Image & 9,385 images & 8 emotional labels, arousal & \\
	\hline \hline
	%MED~\cite{rabiee2016novel} & Video & 31 videos & 8 emotional labels & \\
	
	GECV~\cite{quach2022non} & Video & 627 & \textit{pos} (204), \textit{neu} (221), \textit{neg} (202) & Classification & \textit{train} (90\%), \textit{test} (10\%) \\
	
	\hline
	VGAF~\cite{dhall2020emotiw} & Video & 4,183 & \textit{pos} (1,104), \textit{neu} (1,203), \textit{neg} (1,120) & Classification & \textit{train} (2,661), \textit{val} (766), \textit{test} (756)\\
	
	%\hline
	%GAME-ON~\cite{maman2020game} & Video & 11.5 hours & 6 emotional labels\\
	\hline
	\midrule
	\multicolumn{5}{p{25.14em}}{\textit{pos}: Positive; \textit{neg}: Negative; \textit{neu}: Neutral} \\
	\multicolumn{5}{p{32em}}{\textit{val}: validation.} \\
	\multicolumn{5}{p{50em}}{*For HAPPEI database, we only offer the database volume of each class for train and validation sets.}\\
	%   \multicolumn{4}{p{25.14em}}{\textit{Others} class from CASME II Omitted. } \\
\end{tabular}
\end{table*}

%These methodologies and frameworks have found widespread application in GER. Therefore, this section will elucidate and analyze deep learning-based methods for GER. 

%\textcolor{red}{Figure~\ref{fig:papersurvey} illustrates the  deep learning methods developed in these years.} 

%\textcolor{red}{For GER methodologies, many are founded on features like facial features, gestures, and audio, which exclude background images, and progressively transition from the individual to collective analysis, referred to as bottom-up approaches. Conversely, top-down methods start from the broader context and narrow down to specifics, considering the scene level. Some approaches amalgamate both strategies, forming hybrid methods. Unlike conventional classification techniques, this section predominantly delves into the evolutionary stages of deep learning methods in GER. It  outlines the methodological evolution documented in literature, delineating chronological progressions, database utilization in experiments, and methodological enhancements. Moreover, it furnishes experimental accuracy achieved by certain methodologies across datasets.} 

\section{Input modality}
\label{sec:inputmodality}
Recognizing group-level emotions poses significant challenges due to the diversity in group dynamic, individual emotion expressions, and limited data availability, deep learning (DL) methods, while promising, exhibit varied performance depending on the various modalities utilized. In this section, we describe the diverse information cues based on static image or video sequence utilized for GER and outline their respectively strengths and limitations.

\subsection{Static image}
\label{sec:staticimage}
Due to the abundant availability of facial images online such as AffectNet database~\cite{mollahosseini2017affectnet}, numerous existing studies in FER are conducted on static images. Leveraging the efficiency demonstrated by FER with static images, numerous researchers have extended their efforts to GER, incorporating various additional cues such as face, scene, pose, even objects. Convolutional Neural Networks (CNNs), notably VGG, ResNet and their variants, are commonly employed to analyze static images in GER research.

%\textit{Cues mining.} 

%in social science community should arise group emotions with regards to the pair of a ``top-down approach'' and a ``bottom-up approach''. Therefore, the researcher on GER focused on the hybrid of bottom-up and top-down components. Generally, both bottom-up and top-down are referred to local context and global context, respectively. In other words, Bottom-up is considered to involve the emotion of the individual, while top-down takes into account context such as the background of an image. Considering the benefits of bottom-up and top-down components, some works~\cite{surace2017emotion,balaji2017multi,liu2018enhancing,garg2019group,mou2015group,nagarajan2019group,fujii2020hierarchical,yu2019group,zhu2023towards} analyze multiple inputs to learn features from different cues in group-level images. Specifically, in Garg's work~\cite{garg2019group}, a deep convolutional neural network is used to recognize facial expressions in the image, and a Bayesian network uses scene descriptors to obtain visual features of the image content, thereby inferring the overall emotion of the image. Besides the above modalities, body information~\cite{mou2015group,guo2017group} and object~\cite{yu2019group,zhu2023towards} was incorporated into some works.

\textit{Cue aggregation as input.} Given the flexible nature of group sizes, GER faces the challenge of effectively aggregating features from multiple individuals within a group to derive an overall emotional state.  Existing methods are broadly categorized into two approaches. The first category involves averaging or weighted sum all individuals' emotion scores~\cite{rassadin2017group,lu2019happiness, pan2018group, li2016happiness}, which are typically outputted by a classifier such as Support Vector Machine (SVM). For example, Rassadin~\etal~\cite{rassadin2017group} normalized detected faces and fed them into VGGFace and ImageNet. They constructed an ensemble of four Random Forest classifier trained on the features outputted by VGGFace, ImageNet, and landmark features. Finally, they employed a weighted sum approach to fuse the score outputted by the random forest classifiers. The second approach employs some machine learning algorithms such as bag-of-words and clustering to aggregate all individuals' features into a single feature vector. Balaji~\etal~\cite{balaji2017multi}, for example, utilized CNNs to extract face information. Subsequently, Fisher vector and VLAD encoding techniques were applied to compress all individual features in the image, yielding bottom-up features capable of representing group emotions. This method effectively compresses features, reduces computational complexity. % the bottom-up approach can capture changes in group emotions well.

%Five studies focus on sole modality using facial representation to achieve group emotion estimation. The challenge of such methods lies in how to extract effective feature expressions from the face, and how to effectively integrate the emotional feature expressions of different faces in the group to obtain feature information that conforms to the overall emotional expression.  

%\textit{Kernel-based metrics.} This approach replies on the existing available kernel function to explicitly extract the distance metric or emotion heatmap for GER. For distance metric, two studies~\cite{huang2019analyzing,huang2022group} have focused on the exploration of kernel function. For example,  Huang~\etal~\cite{huang2019analyzing} utilized VGG feature and concentrates on global alignment kernels, designing a support vector machine with the combined global alignment kernels to improve group-level emotion recognition. Additionally, Huang~\etal~\cite{huang2022group} proposed a method based on deep multiple kernel learning based on graph kernels for group-level emotion recognition. For heatmap, one study explored deep learning on an emotion heatmaps. Shamsi~\etal~\cite{shamsi2018group} utilized Gaussian distribution to create an emotional heatmap, which was then used to train CNNs to detect the emotions of a group of people in the image. While emotional heatmaps increase computational complexity, they provide a more intuitive representation of emotional distribution in images, allowing CNNs to extract more effective feature information.

\textit{Multimodality as input.} Group are often considered as ``emotional entities and a rich source of varied manifestations of affect''. Earlier discussions by Bars{\"a}de and Gibson in~\cite{barsade2007does} emphasized the necessity for GER researchers to adopt both a ``top-down approach'' and a ``bottom-up approach''. Consequently, GER research has concentrated on integrating both bottom-up and top-down components. Typically, bottom-up refers to individual emotions, while top-down encompasses contextual factors such as background information in an image. Recognizing the benefits of incorporating both bottom-up and top-down components, several studies~\cite{surace2017emotion,balaji2017multi,liu2018enhancing,garg2019group,mou2015group,nagarajan2019group,fujii2020hierarchical,yu2019group,zhu2023towards} have explored to fuse features from various cues in group-level images. For example, in Garg's work~\cite{garg2019group}, a deep convolutional neural network is utilized to identify facial expressions within an image, while a Bayesian network leverages scene descriptors to extract visual features of the image content, thereby inferring the overall emotion of the image. In addition to these modalities, body information~\cite{mou2015group,guo2017group} and object~\cite{yu2019group,zhu2023towards} have also been incorporated into certain studies. In summary, GER research has commonly explored one or more cues extracted from face, pose/skeleton information, object, and scene context. Cues combination offers the advantage of enabling successful recognition of group-level emotions in the challenging environment when one cue is lacking. However, even employing multiple cues, a key concern arises regarding how to effectively establish connection among these cues to enhance the robustness of GER in real-world scenarios. 

\subsection{Dynamic image sequence}

As emotions are temporal in nature, such automatic tools in environments like factories, companies, and offices can help identifying interventions to maintain a healthy work culture. Facial expressions, being dynamic cues, evolve and change, revealing their effective signals over time. The visual information captured in videos plays a pivotal role in discerning the emotions depicted within them~\cite{krumhuber2023role}. The temporal variations across video frames provides additional information to be exploited, albeit encoding these variations introduces complexity to emotion recognition. Training a network to comprehend the overall affect of a group of people shown across frames poses challenges.  In this subsection, we delineate various dynamic inputs.

\textit{Temporal information as input.} Temporal information encapsulate the dynamics of an entire video sequence in a single instance. The temporal information, modeled by using algorithms like LSTM, has been successfully employed in GER to model scene dynamics and appearance in a video~\cite{sharma2019automatic}. Similarly, active image encapsulated spatial and temporal information from video sequences into a single instance by estimating and accumulating changes in each pixel component. Sun~\etal~\cite{sun2020multi} utilized temporal segment networks to extract RGB information, optical flow frame, and warped optical flow frame for each video, incorporating a temporal shift module to model dynamic scene information. Quach~\etal~\cite{quach2022non} introduced a fusion mechanism called Non Volume Preserving Fusion (NVPF) to better model spatial relationships between facial emotions in each frame, effectively addressing emotional ambiguity caused by insufficient facial resolution or undetectable emotions. 

\textit{Frame aggregation as input.} Dynamic image sequence collected in the wild often include complex scene background. Petrova~\etal~\cite{petrova2020group} designed a method based on VGG-19 framework for each frame, capturing global emotions, followed by using score averaging and accumulation across all frames. Li~\etal~\cite{li2023audio} proposed leveraging multi-task learning theory to aggregate frame features, while Liu~\etal~\cite{liu2020group} employed four aggregation methods including maximum, minimum, average, and standard deviation to consolidate all individual's face feature in a group image. 

\textit{Multimodality as input.} In analyzing group affect, audio features play a crucial role alongside facial image, as relying solely on facial expressions may lead to inaccuracies in estimating overall group affect. Pitch, speech rate, and duration, etc. have been found relevant to affect analysis. In group settings, these features are vital for distinguishing between situations like  arguments and discussions, where the visual model may falter. Visual-audio fusion models have been proposed to enhance visual-based models~\cite{sharma2019automatic,liu2020group,wang2020implicit,pinto2020audiovisual,li2023audio}. For example, Wang~\etal~\cite{wang2020implicit} introduced a network called K-injection audiovisual network, which employs a multi-head cross-attention mechanism to jointly model audio and video data, integrating previous emotion knowledge to improve the model's generalization ability. Recent study indicate that human gesture can convery emotion~\cite{noroozi2018survey}. Several researchers have incorporated human gesture features fused with scene and face cues for video-level GER~\cite{sun2020multi, liu2020group}. For example, Sun~\etal~\cite{sun2020multi} utilized CenterNet for human detection and pose estimation, followed by ResNetSt for extracting body feature. For prediction, the average probability of each frame's prediction was calculated, yielding the video's class prediction. % an audiovisual model, but also introduces a new privacy knowledge injection mechanism from a linguistic and acoustic perspective to incorporate previous emotion knowledge into the audiovisual model, thereby improving the generalization ability. %This method achieved an accuracy of 66.40\% on VAGF, which is 18.9\% higher than Sharma~\etal's method.

\subsection{Discussion}

The primary challenge faced by these methods lies in establishing relationships between modalities and effectively fuse them. As evident from the discussed methods, GER has increasingly emphasized mining the dynamic information present in videos. This trend is expected to drive further advancements in Recurrent Neural Networks (RNN) and its derivatives in group-level emotion recognition. Additionally, these studies have ventured beyong single-modal information, paving the way for research to better comprehend and leverage multimodal information. This broader perspective holds promise for enrich the understanding and utilization of diverse sources of information in group-level emotion recognition tasks.

\section{Deep Networks for GER}

\begin{figure*}[t!]
\centering
\includegraphics[width=\linewidth]{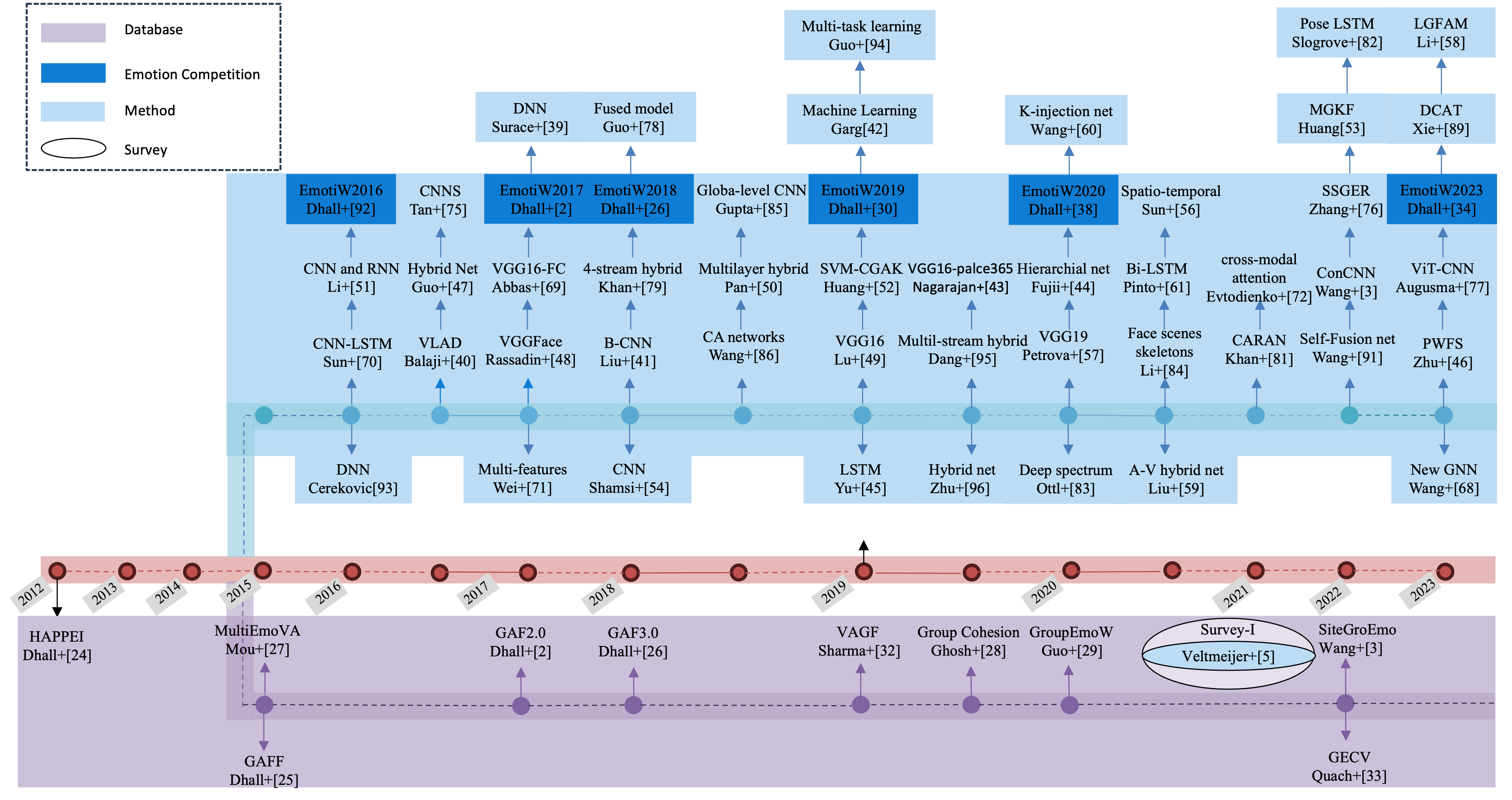}
\caption{The overview of deep learning based technical papers and survey papers for group-level emotion recognition. Viewed in color is BEST.}
\label{fig:papersurvey}
\end{figure*}

Since the inception of artificial neurons in the 1940s, deep learning has been undergone extensive exploration and implementation. Evolving from single-layer perceptrons to multi-layer neural networks, Convolutional neural networks (CNNs), Recurrent neural networks (RNNs), Cascade networks, Graph convolutional networks (GCNs), attention mechanisms, and beyond, deep learning has witnessed rapid evolution. So far, various deep learning approaches have emerged to discern the collective emotions of a group of people, leveraging diverse information such as facial expression, gestures, social interactions, and more. Figure~\ref{fig:papersurvey} illustrates literature spanning database, emotion competition, method, and survey categories of academic papers employing deep learning based methods for GER over the past decade. It is noteworthy that the publication trend has exhibited a noticeable increase, especially attributed to the EmotiW competition. In this section, we delve into the approaches from the perspectives of specialized blocks, network architecture, fusion stage and scheme, and loss function.%Figure~\ref{fig:papertrend} and Figure~\ref{fig:papersurvey} illustrate the annual count and detailed distribution across database, emotion competition, method, and survey categories of academic papers utilizing deep learning based methods for GER over the last decade. It is noteworthy that the publication trend has exhibited a noticeable increase, especially attributed to the EmotiW competition. In this section, we introduced the approaches in the view of special blocks, network architecture, training strategy, and loss.

% \begin{figure}[t!]
% 	\centering
% 	\includegraphics[width=\linewidth]{figures/papertrend.png}
% 	\caption{Number of published papers each year on deep learning for GER. Data is source from Google Scholar and Web of Science (WoS).}
% 	\label{fig:papertrend}
% \end{figure}

\subsection{Basic Network Block}

\begin{figure*}[h]
\centering
\begin{subfigure}[b]{0.45\textwidth}
	\includegraphics[width=0.9\textwidth]{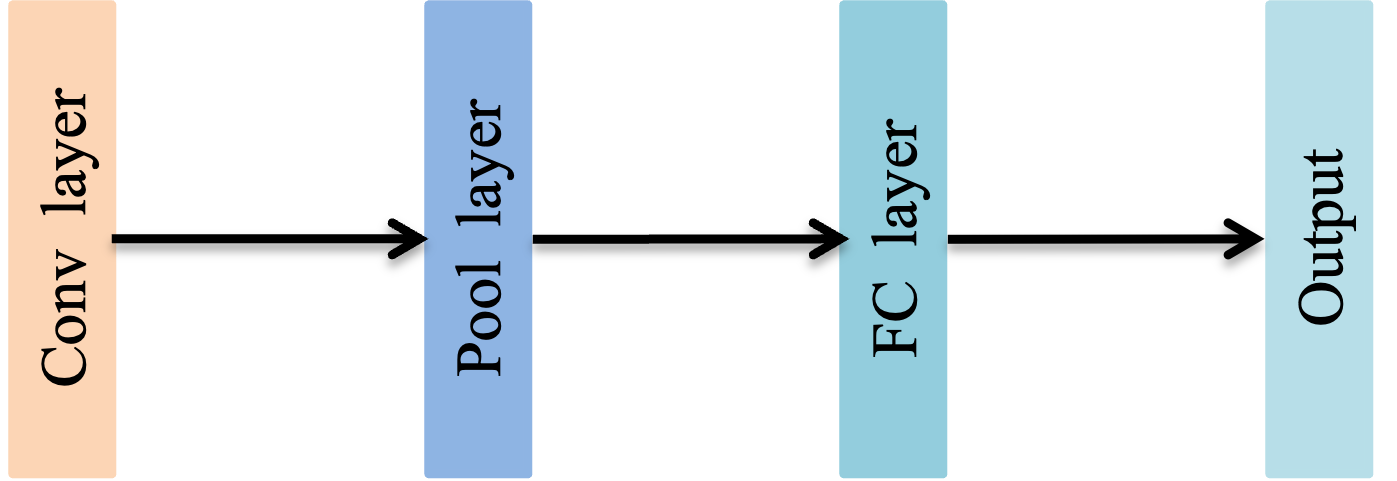}
	\caption{CNN}
	\label{fig:cnn}
\end{subfigure}
\hfill
\begin{subfigure}[b]{0.45\textwidth}
	\includegraphics[width=\textwidth]{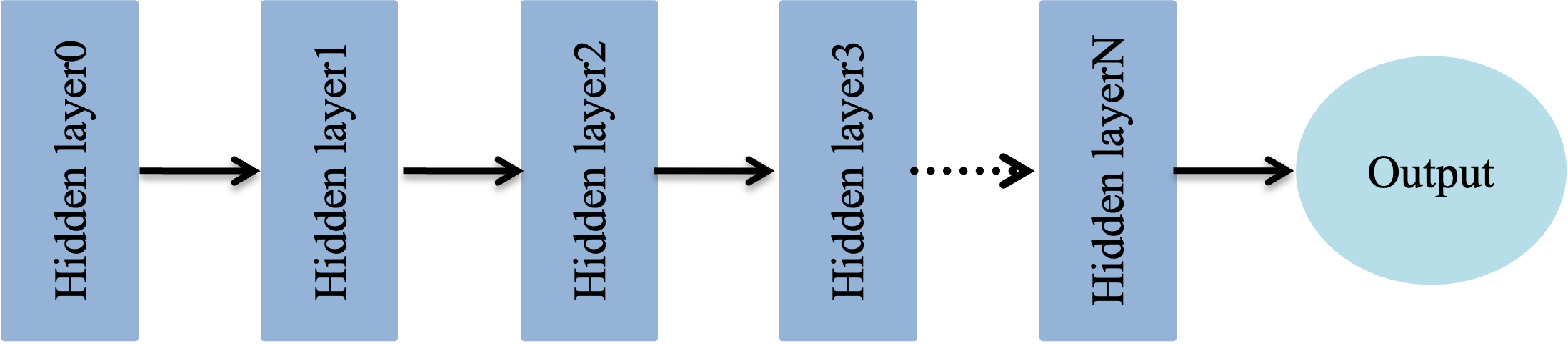}
	\caption{RNN}
	\label{fig:rnn}
\end{subfigure}
\\
\begin{subfigure}[b]{0.45\textwidth}
	\includegraphics[width=\textwidth]{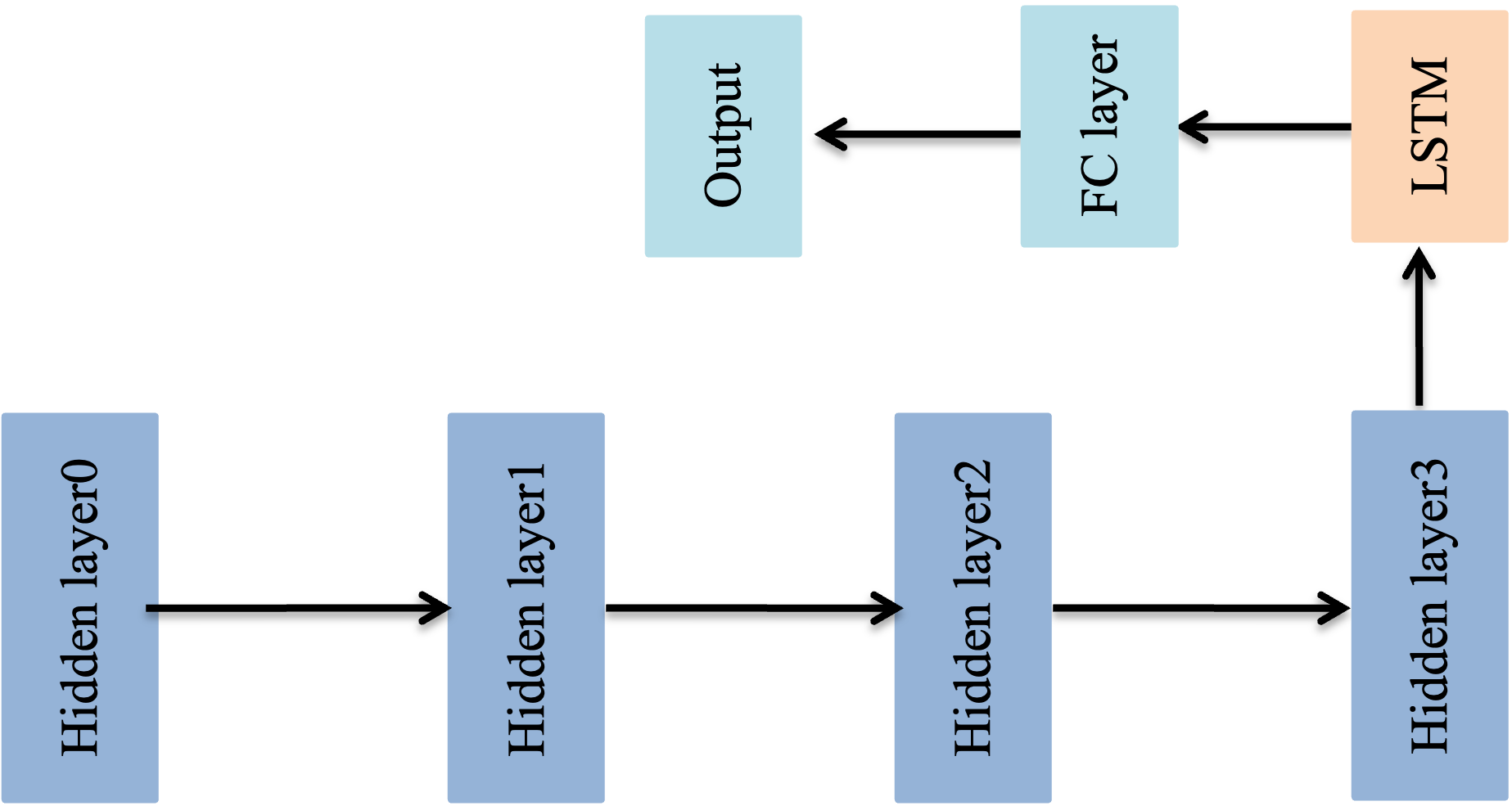}
	\caption{The cascade network}
	\label{fig:cnn_lstm}
\end{subfigure}
\hfill
\begin{subfigure}[b]{0.45\textwidth}
	\includegraphics[width=\textwidth]{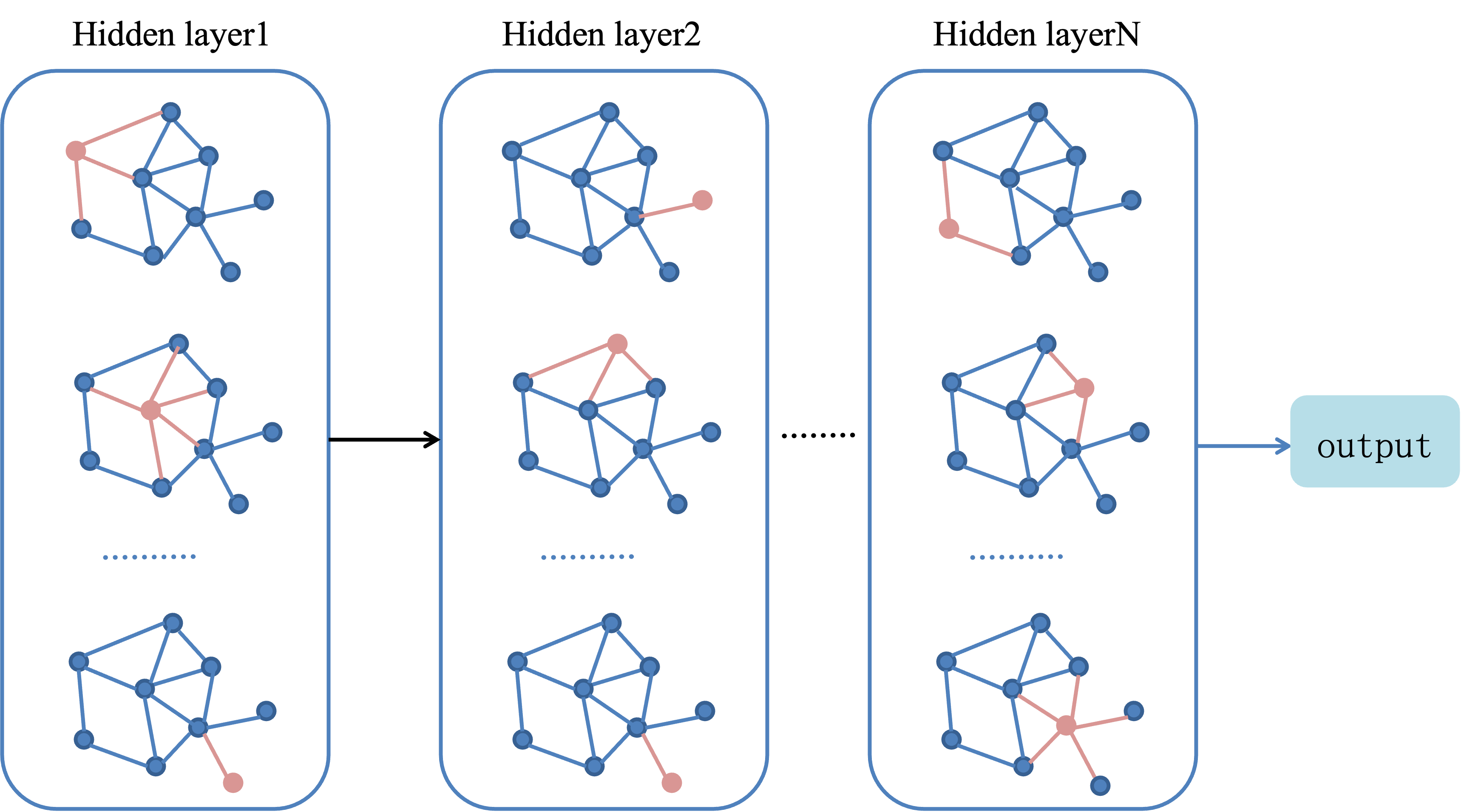}
	\caption{GCN}
	\label{fig:gnn}
\end{subfigure}
\caption{Basic blocks for GER: (1) Convolution block; (b) Recurrent neural network (RNN); (c) Cascade network; (d) Graph Convolutional Network (GCN).}
\label{fig:basicblock}
\end{figure*}

%\textcolor{red}{Lack the block image: CNN, RNN/LSTM, GCN, Cascade network, Attention mechanism, single stream, multiple stream}
%Next, this article will discuss methods of group-level emotion recognition based on convolutional neural networks.

%\textcolor{red}{Furthermore, considering the fact that CNN with more convolutional layers has stronger representation ability, but easy to overfit on small-scale datasets, paper [60] and [169] introduced Recurrent Convolutional Network (RCN) which achieved a shallow architecture though recurrent connec- tions, as shown in Fig. 7 (c).}

%Therefore, this section will discuss various methods based on Recurrent Neural Networks in the field of group-level emotion recognition.

Before describing the network architecture, Figure~\ref{fig:basicblock} first introduce basic network block widely used in GER, including CNN, RNN, GCN. 

\textbf{CNN.} Due to the limitations of traditional machine learning in addressing complex environments, many researchers have explored more in-depth research methods. LeCun~\etal~\cite{lecun1989backpropagation} first proposed a convolutional neural network model based on the backpropagation algorithm. However, due to hardware limitations at the time, the research progress of CNN was relatively slow. It was not until 2012 when Krizhevsky~\etal~\cite{Krizhevsky2012} proposed the AlexNet CNN model in the ImageNet Large Scale Visual Recognition Challenge, which significantly surpassed traditional machine learning methods in accuracy and promoted the development of deep learning in the field of computer vision. Since then, various models based on CNN have been proposed, such as VGGNet, GoogLeNet, ResNet, DenseNet, and MobileNet. Meanwhile, CNN models have also shown their superiority in GER. 

\textbf{RNN.} In Convolutional Neural Networks (CNNs), each input and output are independent of each other, but this ignores the relationship between them. Although CNNs can extract good features when processing image datasets, they are not ideal for datasets with time series data, such as speech, audio, and video. Rumelhart~\etal~\cite{rumelhart1986learning} proposed a method of Recurrent Neural Networks (RNN) that learns and trains through the backpropagation algorithm, and applied it to process data with time series. RNN has the characteristic of introducing a recurrent structure, allowing the network to remember and utilize previous information, thereby extracting better time series features. Various improved network architectures based on RNN have also been widely used, such as Simple Recurrent Neural Network (SRNN), Bidirectional Recurrent Neural Network (BRNN), Long Short-Term Memory Network (LSTM), and Gated Recurrent Unit (GRU), which have achieved good results based on RNN. Normally, RNN is always combined with CNN, leading to the cascade network for GER, as shown in Figure~\ref{fig:cnn_lstm}.

\textbf{GCN.} Graph data has a wide presence in the real world, such as social networks, biological networks, recommendation networks, and chemical molecules. However, previous deep learning models based on CNN and RNN mainly deal with vector and matrix data, which neglects the topological structure of graphs and the relationships between nodes, leading to possible information loss and performance degradation. In order to solve this problem, Scarselli~\etal~\cite{scarselli2008graph} first proposed a new neural network model, namely, the graph neural network model, which extends existing neural network methods to handle data represented in the graph domain. Recent studies demonstrate that the effectiveness of graph convolutional networks (GCNs) in modeling semantic relationships, making them valuable for facial expression recognition (FER) tasks~\cite{Liu2020}, as depicted in Figure~\ref{fig:gnn}. With the success of GCNs in FER, \cite{wang2023implementing} introduced GCNs for GER, aiming to enhance performance by capturing inter-individual relationships.

\subsection{Network architecture}

The efficacy of FER neural units depends on how multiple networks are integrated. GER methods typically adopt one of five network architectures: single-stream, multi-stream, cascade, graph convolutional network, and attention mechanism. In this section, we delve into the specifics of each architecture.
%\textcolor{red}{Deep learning based GER mostly use image-level datasets or video-level dataset and mainly extract individual features, posture features, scene features, global image features, etc. to estimate group emotions.}

\subsubsection{Single-stream networks}

Typical deep GER methods adopt single CNN with individual input. In single-stream 2D CNNs, the primary input is facial images, while single-stream 3D CNNs directly extract spatial and temporal features from video sequences. Many studies~\cite{abbas2017group,rassadin2017group,lu2019happiness} employ transfer learning strategy on deep networks pretrained on large-scale face datasets to mitigate the overfitting issues. For example, Rassadin~\etal~\cite{rassadin2017group} employ a pre-trained VGGFace model on detected faces, followed by a weighted sum to obtain the final result using a random forest classifier. Similarly, Lu~\etal~\cite{lu2019happiness} utilize a VGG model pretrained on the VGGFace dataset to extract facial features for GER. 

In addition to transfer learning methods, several works design cascade network~\cite{sun2016lstm,wei2017new} or kernel methods~\cite{huang2022group} on single-stream shallow CNNs. Sun~\etal~\cite{sun2016lstm} explore various handcrafted feature (LBP), AlexNet, Reduced AlexNet and ResNet, followed by group-expression model or LSTM for group-level happiness intensity estimation. Building on this, Wei~\etal~\cite{wei2017new} extend the group-level intensity estimation with VGGFace pretrained VGG-Face dataset. Furthermore, GER with ResNet18, ResNet34, MobileNet, DenseNet, Resnet50, Inception, GoogleNet, and VGG19 pretrained on Imagenet for scene-level information is explored in~\cite{petrova2020group,pan2018group}. The results demonstrate that VGG surpasses other architectures in GER and excels at distinguishing the complex hidden information in data.

While the aforementioned works are based on 2D CNN with image input, several works employ 3D CNN variants~\cite{pinto2020audiovisual,sun2020multi,sun2020multi} or cascade network~\cite{evtodienko2021multimodal} to directly extract spatial and temporal features from video sequences. Inflated ResNet-3D~\cite{pinto2020audiovisual}, Temporal shift module (TSM)~\cite{sun2020multi}, and Temporal Binding Network (TBN)~\cite{sun2020multi} are introduced in GER. Additionally, a end-to-end cross-attention cascade network~\cite{evtodienko2021multimodal} combined the idea of ClipBERT~\cite{lei2021less} modules with the temporal sequence to enhance the representation in spatial and temporal dimensions.

\subsubsection{\textcolor{black}{Multi-stream Networks}}

Single-stream model represent a basic structure in GER, extracting features solely from a single viewpoint such as the face or scene within group-level image. However, since group-level images encompass diverse and rich information, a single view may not provide sufficient insight. As we discussed in Section~\ref{sec:inputmodality}, employing various inputs from different perspectives can effectively explore spatial and temporal information. Hence, multi-stream networks have been adopted in GER to extract features through multiple inputs. Generally, multi-stream networks can be  categorized into networks with two inputs, more than two inputs, and handcrafted features.

\textit{Multi-stream networks with two inputs.} According to~\cite{bombari2013emotion}, the face plays a crucial role in expressing the emotion. Therefore, among multi-stream networks with local and global inputs, the face remains a primary input. Several studies incorporated the face as local information and the scene as global information. They combined a convolution neural network for face while another neural network on scene descriptors for GER~\cite{li2016happiness, surace2017emotion,abbas2017group,tan2017group,liu2018enhancing}. Experiment results of these studies demonstrate that models based on face outperforms those on other methods and other modality is still competitive and useful to GER. Moreover, the multi-stream networks have shown improvement in GER. Furthermore, Zhang~\etal~\cite{zhang2022semi} presented a semi-supervised group-level emotion recognition (SSGER) framework based on contrastive learning, learning efficient features from both labeled and unlabeled images, where face images and scene images are utilized. In addition to visual features, audio feature are also being considered in dynamic GER. Augusma~\etal~\cite{augusma2023multimodal} proposed branches for both video and audio, with cross-attention between modalities for GER. In their work, the video branch is based on a fine-tuned ViT architecture, while the audio branch extracts Mel-spectrograms and feeds them through CNN blocks into a transformer encoder. 

\textit{Multi-stream networks with more than two inputs.} As mentioned in the previous section, bottom-up components involve individual emotions, while top-down components consider contextual factors such as the background of an image~\cite{barsade2007does}. Therefore, to enhance group feature representation, some works~\cite{guo2017group,liu2020group, guo2018group,khan2018group,fujii2019hierarchical,fujii2020hierarchical,sun2020multi,khan2021regional,wang2023implementing} have investigated combinations of more than one top-down component and bottom-up components. Guo~\etal~\cite{guo2017group} designed a hybrid network that incorporates  scene features, skeleton features, and local facial features with deep convolutional neural networks. Additionally, they~\cite{guo2018group} further used visual attention attention mechanism to fuse face, scene, skeletons, and salient regions. Different from the aforementioned studies, Fujii~\etal~\cite{fujii2019hierarchical,fujii2020hierarchical} proposed a two-stage architecture for GER. The first stage performs binary classification based on facial expression to distinguish “Positive” labels, including discriminative facial expressions from others. For second stage, they considered exploiting object-wise semantic information and scene background for the second classification. Recently, several researchers~\cite{sun2020multi,liu2020group} investigated multi-stream network for dynamic GER. Spatio-temporal features and static features were exploited by Sun~\etal~\cite{sun2020multi}. The fusion of several spatio-temporal modality adopted RGB, RGB difference, optical flow, warped optical flow, and audio, while image-level CNNs were designed based on face and body images. Moreover, a hybrid network fusing audio stream, facial emotion stream, environmental object statistics stream (EOS), and video stream are designed with temporal shift module and SVM for GER~\cite{liu2020group}. 

\textit{Multi-stream networks with handcafted features.} According to the analysis, the facial emotion or movements of group-level emotion are highly related to face textures, while scene information contains more abundant information for GER, the handcrafted features for low-level representation also plays an important role in GER. Multiple works~\cite{wei2017new,balaji2017multi} combined deep features for face-level and handcrafted features for scene-level to leverage the low-level and high-level information for robust GER.

\subsubsection{Cascade network}

For GER, dealing with varying numbers of face between group-level images poses a significant challenge. As discussed in Section~\ref{sec:staticimage}, current cascade networks can be categorized to two types. The first category utilizes variants of CNN such as ResNet, VGG, followed by decision-level score fusion. In contrast, the second category combines various CNN and RNNs to address the inconsistency in the number of faces between two group-level image. In this section, we will discuss the details of the second category~\cite{sun2016lstm,guo2018group,slogrove2022group,evtodienko2021multimodal,ottl2020group}.

%. As discussed in Section~\ref{sec:staticimage}, current cascade networks can be are not always consistent between two images. In other words, two images contain different number of faces. It is tough to directly use a sole network to classify the group-level emotion. Therefore, the current cascade networks can be classified to two categories. The first category is to use a variant of CNN such as ResNet, VGG, followed by decision-level score fusion. It was described in Section~\ref{sec:staticimage}. The other one is to combine various CNN and RNN to address the inconsistent number of face between two group-level image. In this section, we will discuss the details of the second category~\cite{sun2016lstm,guo2018group,slogrove2022group,evtodienko2021multimodal,ottl2020group}.

In GER, a critical issue is how to effectively model the variability in the number of faces. LSTM is a primary method. Sun~\etal~\cite{sun2016lstm} were the first to investigate the combination of CNNs and LSTM. They explored the use of AlexNet, Reduced AlexNet, and ResNet to extract the individual faces in a group. Subsequently, a weighted LSTM was used to assign different weights to each facial feature based on factors such as the size of the face and the distance between them. Furthermore, visual attention mechanisms were incorporated in LSTM for GER~\cite{guo2018group, evtodienko2021multimodal}. In~\cite{evtodienko2021multimodal}, a cascade network containing CNN and LSTM was employed to extract image-level and audio-level features, respectively. An attention mechanism was then introduced to compute important features at each time step. Additionally, Li~\etal~\cite{li2020group} employed a similar architecture at the face-level for GER. However, in contrast to~\cite{sun2016lstm}, skeletons and scene features were directly fused at the end. Moreover, skeletons information extracted by OpenPose toolkit was considered in cascade network proposed by Slogrove~\etal~\cite{slogrove2022group}. In their approach, the coordinates and confidence information of all individual key points were fed into an LSTM as a sequence for modeling, resulting in a group-level emotion classification. Additionally, speech signals were investigated for GER in~\cite{ottl2020group}. They proposed a method incorporating a cascade network with multi-task learning for GER, using deep spectral features on speech signal. The cascade network based on CNN and RNN, was designed to extract discriminative deep spectral features, while multi-task learning combines emotion recognition and speaker identification tasks during the model training process.

\subsubsection{GCN based network}
Individuals within a group often exhibit diverse social relationships with others. To highlight this social aspect, several works~\cite{guo2020graph, wang2022congnn} have utilized graph convolutional networks (GCNs) to model both visual features and social context within group-level images. In these approaches, the emotional states of individuals are treated as node features, while the interactions between individuals are represented as graph edges, thus forming a graph structure. Guo~\etal~\cite{guo2020graph} introduced a group-level emotion recognition method based on four cues, where faces, bodies, objects, and the entire image are transformed into a graph structure. This graph represents the relationships within the group based on these four cues, facilitating group-level emotion recognition. Additionally, Wang~\etal~\cite{wang2022congnn} proposed a context-consistent cross-graph neural network to mitigate emotional biases resulting from different cues in multi-cue emotion recognition.

%An individual may exhibit diverse social relationship between other individual in a group. In order to emphasize the social feature, several works~\cite{guo2020graph,wang2022congnn} have exploits graph convolutional network (GCN) to model the visual features and social context within a group-level image. The emotional state of individuals are viewed as node features, while the interaction between individuals as graph edges, thereby constructing a graph structure. Guo~\etal~\cite{guo2020graph} proposed a group-level emotion recognition method based on four clues, mainly transforming faces, bodies, objects, and the entire image into a graph structure, and constructing a graph to represent the relationships within the group based on these four clues, in order to recognize group emotion. Furthermore, a context consistent cross graph neural network is  proposed to alleviate the emotional bias caused by different clues in multi clue emotion recognition~\cite{wang2022congnn}.

\subsubsection{Attention based network}

In order to prioritize important character or object features that play a pivotal role in group emotions, five studies have incorporated attention mechanisms. Gupta~\etal~\cite{gupta2018attention} detect local facial emotions by employing an attention mechanism to concentrate on more pertinent local information on the face, generating probability attention weights through the Softmax function. The weighted sum of facial features is then calculated based on these attention weights to produce a single facial feature vector representation. Additionall, Guo~\etal~\cite{guo2018group} and Khan~\etal~\cite{khan2021regional} also integrated attention mechanisms. They introduced a novel region attention network (RAN) to detect and extract crucial features in facial regions. The region attention mechanism comprises an attention generator and an attention applier, where the former generates adaptive region attention weights to emphasize important facial features in different regions, while the latter applies these generated region attention weights to the output of the feature extractor to enhance the distinction of facial features.  Furthermore, Wang~\etal~\cite{wang2018cascade} proposed a cascaded attention network, which leverages the importance of each face in the image to generate a global representation based on all faces, effectively focusing on the feature information of the most important face.

%In order to focus on important character or object features that play a key role in group emotions, five studies have introduced attention mechanisms. Gupta~\etal~detect local facial emotions using an attention mechanism to focus on mor e useful local information on the face, and generate probability attention weights through the Softmax function~\cite{gupta2018attention}. The weighted sum of facial features is calculated based on the attention weights to generate a single facial feature vector representation. Additionally, Guo~\etal~\cite{guo2018group} and Khan~\etal~\cite{khan2021regional} also used attention mechanisms. They proposed a new region attention network (RAN) to detect and extract important features in facial regions. The region attention mechanism includes an attention generator and an attention applier, where the former is used to generate adaptive region attention weights to focus on important facial features in different regions. The latter applies the generated region attention weights to the output of the feature extractor to improve the distinction of facial features. Furthermore, Wang~\etal~\cite{wang2018cascade} proposed a cascaded attention network, which uses the importance of each face in the image to generate a global representation based on all faces. This method can effectively focus on the feature information of the most important face.

Moreover, Transformer-based architectures have found extensive applications in NLP~\cite{vaswani2017attention} and computer vision~\cite{han2022survey} tasks. Inspired by the significant success of Transformer architecture in various tasks, two recent studies~\cite{augusma2023multimodal,xie2023most} have explored the application of transformers to the GER task. Augusma et al.~\cite{augusma2023multimodal} initially utilized the visual Transformer mechanism and the BERT framework to extract features from global images and speech, respectively. They employed a cross-attention mechanism to learn the weights of the two modes and subsequently performed weight fusion. Moreover, a dual-branch cross-patch attention Transformer (DCAT) was proposed to incorporate the psychological concept of the Most Important Person (MIP) and the global image.

%Moreover, Transformer-based architecture have been extensively in NLP~\cite{vaswani2017attention} and computer vision~\cite{han2022survey} tasks. Inspired by the significant success of Transformer architecture in various tasks, two studies~\cite{augusma2023multimodal,xie2023most} have recently discussed the application of transformers to GER task. Augusta~\etal~\cite{augusma2023multimodal} first used the visual Transformer mechanism and the BERT framework to extract features from global images and speech, respectively. Meanwhile, they use a cross attention mechanism to learn the weights of the two modes, and finally performs weight fusion. Moreover, a dual branch cross patch attention Transformer (DCAT) was proposed to incorporate the psychological concept of the Most Important Person (MIP) and global image.

In summary, GER network architectures can be broadly categorized into single-stream, multi-stream, cascade networks, GNN-based networks, and attention-based networks. While single-stream serves as the fundamental model, it only considers a single view of the group-level image. To leverage more information, multi-stream networks learn features from multiple perspectives for robust GER. Additionally, as the group size fluctuates, cascade networks sequentially incorporate various modules like RNNs and LSTMs to construct an end-to-end GER network. GNN effectively models interactions between individuals based on social relationships. Conversely, the attention mechanism, inspired by the psychological concept of the most important person, focuses on extracting key features from all individuals. In the future, combining more effective modules in multi-stream, cascade, and attention-based approaches could further enhance GER performance.

\subsection{Fusion stage and scheme}

In GER, the fusion stage plays a crucial role in integrating information from multiple cues to improve the accuracy of emotion recognition. It encompass various methods for combining features extracted from different modalities, such as facial expressions, scene, skeleton etc. One common fusion approach is score-level, as used in studies like ~\cite{guo2018group, yu2019group, nagarajan2019group}. In their approaches, the output scores from individual modalities are combined using techniques like averaging or mean voting.

Alternatively, feature-level fusion used in studies like those by~\cite{fujii2019hierarchical, fujii2020hierarchical, quach2022non}, integrates raw feature representations extracted from each modality before feeding them into a classifier. This approach allows the model to learn more complex relationships between different modalities but may be more computationally intensive. Besides score-level and feature-level fusion, kernel-based~\cite{huang2019analyzing, huang2022group} and loss function-based~\cite{zhu2023towards, zhang2022semi} fusion are two alternative ways to fuse multi-modality. For example,~\cite{zhang2022semi} proposed a weight cross-entropy loss function on Scene-Face network by combining face and scene information.

\subsection{Loss function}

Different from classical methods, where the feature extraction and classification are independent, deep networks can perform end-to-end classification through loss functions by penalizing the deviation between predicted and true labels during training. Most GER works directly apply the commonly used softmax cross-entropy loss~\cite{rassadin2017group}. The softmax loss is typically effective at correctly classifying known categories. However, in practical classification tasks, the classification of unknown samples is also essential. Therefore, to achieve better generalization ability, it is crucial to further enhance inter-class difference and reduce intra-class variation, especailly for data scarcity. Metric learning techniques, such as contrastive loss~\cite{wang2021understanding}, have been developed to ensure intra-class compactness and inter-class separability by measuring the relative distances between inputs.  Wang~\etal~\cite{wang2022self} proposed a contrastive learning-based self-attentive network. In this approach, different features are embedded into a vector space, and the similarity and difference of features are learned by enhancing the similarity between samples of the same class and reducing the similarity between samples of different classes. Then, adaptive weight calculation and weighted average fusion are performed to adaptively fuse features at different levels. Although the above two methods have achieved good performance, they are still limited to static images. Additionally, metric learning loss often requires effective sample mining strategies for robust recognition performance. Metric learning alone may not suffice for learning a discriminative metric space for GER. To address these challenges, Zhang~\etal~\cite{zhang2022semi} proposed a semi-supervised group-level emotion recognition framework based on contrastive learning to learn efficient features from both labeled and unlabeled images. To alleviate the uncertainty of given pseudo-labels, they introduce Weight Cross-Entropy Loss (WCE-Loss) to suppress the influence of samples with unreliable pseudo-labels in the training process. 

In summary, although most current GER approaches utilize the standard softmax cross-entropy loss, only a limited number of studies have explored alternative loss functions such as contrastive learning loss or introduced novel loss functions to enhance inter-class separability, intra-class compactness, and achieve well-balanced learning. Looking ahead, investigating more effective loss functions targeting discriminative representations for group-level emotion features holds significant promise as a direction for future research in GER.

%\textcolor{red}{In summary, most current GER approaches are based on the basic softmax cross-entropy loss. Few studies utilized the contrastive learning loss or proposed a new loss function to encourage inter-class separability, intra-class compactness, and balanced learning. In the future, exploring more effective loss functions to learn discriminative representation for group-level feature can be a promising research direction.}

%\subsection{\textcolor{red}{Discussion}}

\section{\textcolor{black}{Experiments}}

\subsection{Performance metric}

The standard evaluation metric for GER typically involves using accuracy for group-level emotion recognition and mean square error for group happiness estimation. Accuracy assesses the proportion of correct predictions relative to the total number of evaluated samples, providing a measure of the model's overall performance in correctly identifying group-level emotion. Conversely, mean square error quantifies the average squared difference between the predicted and true happiness values, offering a measure of the model's accuracy in estimating the happiness levels of a group.

\subsection{Model Evaluation Protocols}

Cross-validation stands as a widely utilized protocol for evaluating GER performance. This protocol involves dividing the dataset into train, validation, and test sets, ensuring fair verification of deep learning architectures on group emotion datasets. Cross-validation in the GER contains fixed partition validation and K-fold cross validation. The first kind of cross-validation is commonly used as the official evaluation method in the competitions such as the Emotion Recognition in the wild  challenge (EmotiW)~\cite{dhall2018emotiw}. Here, the train, validation, and test sets are pre-determined and remain fixed throughout the competition, eliminating randomization. Participants receive the train and validation sets at the competition's outset for model development, while the test set is revealed later for final performance assessment and ranking. On the other hand, K-fold cross-validation protocol is also prevalent in GER research. This protocol involves randomly dividing the dataset into $k$ equally sized parts, with each part serving as a test set in turn while the remaining portions constitute the training data. The process repeats $k$ times, with each partition serving as the test set once. The choice for $k$, typically 4 or 5 in GER, can significantly impact evaluating time while ensuring robust performance assessment.

%Competition protocol in EmotiW~\cite{dhall2018emotiw}: The fixed partition validation is commonly used as an official protocol in the series of Emotion Recognition in the Wild (EmotiW) challenge competitions. In the official protocol, the train, validation, and test sets are strictly fixed without random validation. The train and validation sets were released for the model design when the competition started, while the test set is finally released for ranking all teams at the end of the competition. As presented in Table~\ref{tab:databaseintro}, the K-fold cross validation protocol is also employed in GER. In K-fold cross-validation, the original samples are randomly divided into k equally sized parts. Each part is used as a test set in turn, while the remaining parts constitute the training data. Consequently, the cross-validation process is repeated $K$ times. In practice, selecting an appropriate value for $K$ can significantly reduce evaluation time. For GER, typical choice for $K$ include 4 or 5. 

\subsection{Performance analysis }

Table~\ref{tab:EmotiW} reports the performance of various deep learning models for GER as reported in EmotiW since 2016. With the exceptions of~\cite{petrova2020group,liu2020group}, most methods proposed in these competition aim to fuse more than two cues such as face, scene, and skeleton, among others. Notably, established neural network architectures such as VGG have been commonly employed for extracting facial expression feature. Additionally, to accommodate the variable number of faces/objects, variations of RNN have been prevalent in many algorithms. Furthermore, since 2019, attention modules, facilitated by the successful application of transformer architectures, have gained widespread adoption in Emotion Recognition in the Wild competitions. Regarding performance metrics, in the HAPPEI dataset, the lowest reported RMSE is 0.822. In the GAF2.0 competition, the highest reported accuracy is 80.9\%, while in GAF3.0, it is 68.08\%. This suggests that GAF3.0 introduced more competitive samples and increased the challenge level. Additionally, a new track called Group Cohesion, derived from GAF3.0, was introduced to evaluate group cohesion. Despite improvements in team performance between EmotionW2016 and EmotiW2019, there is still significant room for enhancing overall performance. Since 2020, the utilization of video-based datasets has become prevalent, indicating a growing consideration for temporal information in algorithm development. Moreover, traditional feature extraction techniques have been gradually supplanted by deep learning methods in recent years. 
%with the successful application of transformer architecture, attention module has been widely in Emotion Recognition in the Wild since 2019. In the performance, in HAPPEI, the best RMSE is 0.822. In GAF2.0, the best performance in the competion is 80.9\%, while in GAF3.0 68.08\%. It demonstrates that GAF3.0 added more competitive samples and more challenge. Group cohesion is a new track derived from GAF3.0 to evaluate the cohesion of group. Although the performance of team in EmotiW2019 is better than EmotiW2016, it has still much space to improve the performance. Since 2020, the video-based database is firstly applied. It is seen that temporal information is considered in the algorithms. Additionally, traditional feature extraction is also included. With temporal information, the accuracy is further improved comparing with EmotiW2018. 

Beyond competition, many researchers have also explored GER using group-level emotion databases like HAPPEI and GAF. Tables~\ref{tab:method} and~\ref{tab:VGAF} offer detailed comparison of method or model accuracies proposed by researchers over the last decade, excluding those participating in EmotionW competition. Notably, except for~\cite{shamsi2018group, huang2022group, huang2019analyzing, quach2022non}, all methods fused more than two cues. Among these, face, pose/skeleton, and object are regarded as local component, while scene information serves as global information. This approach aligns with the concept of bottom-up and top-down components mentioned in group emotion theory. Furthermore, widely recognized network blocks such as VGG, Xception, ResNet and AlexNet have found extensive use in GER due to their promising performance in image and face recognition tasks.  Additionally, the fusion of various networks enables better exploitation of complementary information between them. Common fusion schemes include Average and Feature concatenation, which provide straightforward solutions to the fusion problem. However, some researchers have proposed novel approaches to fuse multi-modality features. For example, Guo~\etal~\cite{guo2020graph} introduced a graph convolutional network to facilitate information exchange among features extracted from different models. Zhu~\etal~\cite{zhu2023towards} proposed a uncertain-aware learning to extract more robust representation from face, object, and scene modalities for GER. Moreover, unlike competitions, recently proposed GER methods have been evaluated across various databases, such as GroupEmoW, GAF, and GECV-GroupImg to access their generalization ability. With the introduction of databases like VGAF and GECV-GroupVid, researchers have started exploring the spatiotemporal GER.

In general, modality fusion can yield promising results across all datasets. Different modalities contribute diverse information, allowing for more comprehensive exploration of limited GER samples. Since the combined inputs offer robust GER solutions, multi-stream networks are recommended to effectively learn representations from available modalities. In contrast, single-modality approaches perform worse due to limited information and redundancy.

From Tables~\ref{tab:method} and~\ref{tab:VGAF}, it is clear that fusion of scores and features is a common approach in integrating multiple modalities. Additionally, there is a growing trend towards using loss functions for multi-modality fusion. Fusion schemes like cross-attention, GCN, ECL, and NVPF have demonstrated state-of-the-art results across all databases. This is likely due to the challenges posed bythat the limited number of GER samples and group sizes, making leveraging additional data sources as reasonable and effective solution.

Presently, scene information based on LSTM and averaging features across all faces are commonly employed in video-based GER studies. This is likely because flexible group sizes pose the main challenge for video-based GER. Recently, Quach~\etal~\cite{quach2022non} proposed a Non-Volume Preserving Fusion (NVPF) mechanism with LSTM to model spatial representation between groups of multiple faces and temporal relationships between multiple video frames. However, the small-sample GE dataset limits the ability to model group-level features for video-based GER. The combination of transfer learning and graph-based methods is anticipated to be a promising direction for future GER studies.

\begin{table*}[th!]
% \begin{sidewaystable}
	\centering
	\caption{Performance comparison of remarkable deep learning techniques published in ACM Library in Group emotion competition in Emotion Recognition in the Wild (EmotiW) challenge~\cite{dhall2016emotiw, dhall2017individual,dhall2018emotiw,dhall2019emotiw,dhall2020emotiw,dhall2023emotiw}. The best performance for specific database on the test set is bold and red color. }
	\vspace{10pt}
	\label{tab:EmotiW}
	\scriptsize
	\begin{tabular}{|c|c|c|ccccc|c|c|c|c|c|c|}
		\hline
		\multirow{2}{*}{\makecell[c]{Dataset\\(Year)}} & \multirow{2}{*}{Cate.} & \multirow{2}{*}{Ref.} & \multicolumn{5}{c|}{Modality} & \multirow{2}{*}{\makecell[c]{Network \\architecture}} & \multirow{2}{*}{\makecell[c]{Fusion \\scheme}} & \multirow{2}{*}{\makecell[c]{Fusion \\stage}} & \multirow{2}{*}{Pre-train}  & \multirow{2}{*}{Prot} & \multirow{2}{*}{Perf.} \\ \cline{4-8}
		&   &  & F & S & P & A & T &    &   &   &  &  &   \\ \hline
		
		\multirow{6}{*}{\makecell[c]{HAPPEI*\\(2016)}} & \multirow{6}{*}{6} & \multirow{2}{*}{\cite{li2016happiness}} & \multirow{2}{*}{\checkmark}  & \multirow{2}{*}{\checkmark} &  \multirow{2}{*}{} & \multirow{2}{*}{} & \multirow{2}{*}{} & \multirow{2}{*}{\makecell[c]{ResNet for F\\CENTRIST+PCA for S}} & \multirow{2}{*}{LSTM} & \multirow{2}{*}{Feature} & \multirow{2}{*}{FER2013} & \textit{val} & 0.494 \\ \cline{13-14}   
		& & & & & & & &  & & & & \textit{test} & \textcolor{red}{\textbf{0.822}} \\ \cline{3-14}
		
		& & \multirow{2}{*}{\cite{sun2016lstm}} & \multirow{2}{*}{\checkmark}  & \multirow{2}{*}{} &  \multirow{2}{*}{} & \multirow{2}{*}{} & \multirow{2}{*}{} & \multirow{2}{*}{\makecell[c]{AlexNet}} & \multirow{2}{*}{\makecell[c]{LSTM}} & \multirow{2}{*}{Feature} & \multirow{2}{*}{FER2013} & \textit{val} &  0.4942 \\ \cline{13-14}   
		& & & & & & & &  & & & & \textit{test} & 0.836 \\ \cline{3-14}
		
		&  & \multirow{2}{*}{\cite{cerekovic2016deep}} & \multirow{2}{*}{\checkmark}  & \multirow{2}{*}{\checkmark} &  \multirow{2}{*}{} & \multirow{2}{*}{} & \multirow{2}{*}{} & \multirow{2}{*}{\makecell[c]{ResNet+LSTM for F\\CENTRIST/VGG for S}} & \multirow{2}{*}{Concat} & \multirow{2}{*}{Feature} & \multirow{2}{*}{-} & \textit{val} & 0.55\\ \cline{13-14}   
		& & & & & & & &  & & && \textit{test} & 0.865 \\  
		\hline
		
		% GAF2.0
		\multirow{14}{*}{\makecell[c]{GAF2.0\\(2017)}} & \multirow{14}{*}{3} & \multirow{2}{*}{\cite{rassadin2017group}} & \multirow{2}{*}{\checkmark}  & \multirow{2}{*}{\checkmark} &  \multirow{2}{*}{} & \multirow{2}{*}{} & \multirow{2}{*}{} & \multirow{2}{*}{\makecell[c]{VGG for F\\ImageNet for S}} & \multirow{2}{*}{\makecell[c]{Soft aggregation\\\&weighting }} & \multirow{2}{*}{Score}& \multirow{2}{*}{\makecell[c]{VGG face\\ ImageNet}} & \textit{val} &  75.39\%  \\ \cline{13-14}   
		& & & & & & & &  & & && \textit{test} & 78.53\% \\ \cline{3-14}
		
		&  & \multirow{2}{*}{\cite{balaji2017multi}} & \multirow{2}{*}{\checkmark}  & \multirow{2}{*}{\checkmark} &  \multirow{2}{*}{} & \multirow{2}{*}{} & \multirow{2}{*}{} & \multirow{2}{*}{\makecell[c]{HOG+FV for S\\VGG+VLAD for F}} & \multirow{2}{*}{Cont} & \multirow{2}{*}{Feature} &\multirow{2}{*}{-} & \textit{val} & 65\%\\ \cline{13-14}   
		& & & & & & & &  & & &&  \textit{test} & 75.10\% \\ \cline{3-14} 
		
		&  & \multirow{2}{*}{\cite{surace2017emotion}} & \multirow{2}{*}{\checkmark}  & \multirow{2}{*}{\checkmark} &  \multirow{2}{*}{} & \multirow{2}{*}{} & \multirow{2}{*}{} & \multirow{2}{*}{\makecell[c]{AlexNet for F\\Context for S}} & \multirow{2}{*}{Bayesian Network} & \multirow{2}{*}{Score} & \multirow{2}{*}{-} & \textit{val} &  67.75\%  \\ \cline{13-14}   
		& & & & & & & &  & & &&  \textit{test} & 64.68\% \\ \cline{3-14}
		
		&  & \multirow{3}{*}{\cite{guo2017group}} & \multirow{3}{*}{\checkmark} & \multirow{3}{*}{\checkmark} &  \multirow{3}{*}{\checkmark} & \multirow{3}{*}{} & \multirow{3}{*}{} & \multirow{3}{*}{\makecell[c]{VGG for F\\ Inception\&ResNet for P\\Inception\&VGG for S}} & \multirow{3}{*}{SVM} & \multirow{3}{*}{Score} & \multirow{3}{*}{\makecell[c]{FER2013\\GENKI-4K}} & \textit{val}&  80.05\%  \\ \cline{13-14}   %ImageNet\\
		& & & & & & & &  & & && \multirow{2}{*}{\textit{test}} & \multirow{2}{*}{\textcolor{red}{\textbf{80.61\%}}} \\ 
		& & & & & & & &  & & && &  \\ \cline{3-14} 
		
		&  & \multirow{2}{*}{\cite{abbas2017group}} & \multirow{2}{*}{\checkmark}  & \multirow{2}{*}{\checkmark} &  \multirow{2}{*}{} & \multirow{2}{*}{} & \multirow{2}{*}{} & \multirow{2}{*}{\makecell[c]{Xception for F\\VGG for S}} & \multirow{2}{*}{Cont} & \multirow{2}{*}{Feature}& \multirow{2}{*}{FER2013} & \textit{val} &  72.38\%  \\ \cline{13-14}   
		& & & & & & & &  & & && \textit{test} & 63.43\% \\ \cline{3-14}
		
		&  & \multirow{2}{*}{\cite{wei2017new}} & \multirow{2}{*}{\checkmark}  & \multirow{2}{*}{\checkmark} &  \multirow{2}{*}{} & \multirow{2}{*}{} & \multirow{2}{*}{} & \multirow{2}{*}{\makecell[c]{VGG+DCNN for F\\CENTRIST+VGG for S}} & \multirow{2}{*}{LSTM/SVM} & \multirow{2}{*}{Feature} & \multirow{2}{*}{-} & \textit{val} &  -  \\ \cline{13-14}   
		& & & & & & & &  & & & & \textit{test} & 79.78\%\\ \cline{3-14}

		&  & \multirow{2}{*}{\cite{tan2017group}} & \multirow{2}{*}{\checkmark}  & \multirow{2}{*}{\checkmark} &  \multirow{2}{*}{} & \multirow{2}{*}{} & \multirow{2}{*}{} & \multirow{2}{*}{\makecell[c]{4-layer CNN for F\\ResNet for S}} & \multirow{2}{*}{Average} & \multirow{2}{*}{Score}& \multirow{2}{*}{\makecell[c]{FERPlus\\Places}} & \textit{val} &  83.7\%  \\ \cline{13-14}     
		& & & & & & & &  & & && \textit{test} & 80.9\% \\ \hline
		
		%% GAF3.0
		\multirow{9}{*}{\makecell[c]{GAF3.0\\(2018)}} & \multirow{9}{*}{3} & \multirow{2}{*}{\cite{guo2018group}} & \multirow{3}{*}{\checkmark}  & \multirow{3}{*}{\checkmark} &  \multirow{3}{*}{\checkmark} & \multirow{3}{*}{} & \multirow{3}{*}{} & \multirow{3}{*}{\makecell[c]{VGG for F\\ Inception/SE-ResNet for S\\ ResNet for P}} & \multirow{2}{*}{\makecell[c]{Weight \\Average}} & \multirow{3}{*}{Score}& \multirow{3}{*}{\makecell[c]{FER2013\\GENKI-4K}} & \textit{val} &  78.98\%  \\ \cline{13-14}    % DL
		& & & & & & & &  & & &&  \multirow{2}{*}{\textit{test}} & \multirow{2}{*}{\textcolor{red}{\textbf{68.08\%}}}\\ 
		&  & & & & & & &  & & &&  &  \\ \cline{3-14}
		
		& &  \multirow{2}{*}{\cite{khan2018group}} & \multirow{2}{*}{\checkmark}  & \multirow{2}{*}{\checkmark} &  \multirow{2}{*}{} & \multirow{2}{*}{} & \multirow{2}{*}{} & \multirow{2}{*}{\makecell[c]{ResNet for F\\VGG for S}} & \multirow{2}{*}{\makecell[c]{Weight\\average}} & \multirow{2}{*}{Score} & \multirow{2}{*}{\makecell[c]{FER2013\\RAF-DB}} & \textit{val} &  78.39\%  \\ \cline{13-14}   % , ImageNet
		& & & & & & & &  & & && \textit{test} & 65.59\% \\  \cline{3-14}
		
		& &  \multirow{2}{*}{\cite{gupta2018attention}} & \multirow{2}{*}{\checkmark}  & \multirow{2}{*}{\checkmark} &  \multirow{2}{*}{\checkmark} & \multirow{2}{*}{} & \multirow{2}{*}{} & \multirow{2}{*}{\makecell[c]{DenseNet for S\\SphereFace for F}} & \multirow{2}{*}{Cont} & \multirow{2}{*}{Feature} &\multirow{2}{*}{\makecell[c]{ImageNet\\CASIA-Webface}} & \textit{val} & 80.98\%\\ \cline{13-14}   
		& & & & & & & &  & & && \textit{test} & 64.83\%  \\ \cline{3-14} 
		
		& &  \multirow{2}{*}{\cite{wang2018cascade}} & \multirow{2}{*}{\checkmark}  & \multirow{2}{*}{\checkmark} &  \multirow{2}{*}{} & \multirow{2}{*}{} & \multirow{2}{*}{} & \multirow{2}{*}{\makecell[c]{CAN for F, ResNet for \\S, SE-net for P}} & \multirow{2}{*}{Average} & \multirow{2}{*}{Score} & \multirow{2}{*}{FERPlus} & \textit{val} &  86.7\%  \\ \cline{13-14}   
		& & & & & & & &  & & &&  \textit{test} & 67.48\% \\ 
		
		\hline 
		
		%% group cohesion
		\multirow{8}{*}{\makecell[c]{Group \\ Cohesion* \\ (2019)}} & \multirow{8}{*}{4} & \multirow{2}{*}{\cite{guo2019exploring}} & \multirow{2}{*}{\checkmark}  & \multirow{2}{*}{\checkmark} &  \multirow{2}{*}{\checkmark} & \multirow{2}{*}{} & \multirow{2}{*}{} & \multirow{2}{*}{\makecell[c]{CAN for F\\SE-Net for S/P}} & \multirow{2}{*}{Average} & \multirow{2}{*}{Score} & \multirow{2}{*}{\makecell[c]{FERPlus\\ImageNet}} &  \textit{val} & 0.5588\\ \cline{13-14}   
		& & & & & & & &  & & &&  \textit{test} & 0.4382 \\ \cline{3-14}
		
		&  &\multirow{3}{*}{\cite{xuan2019group}} & \multirow{3}{*}{\checkmark}  & \multirow{3}{*}{\checkmark} &  \multirow{3}{*}{\checkmark} & \multirow{3}{*}{} & \multirow{3}{*}{} & \multirow{3}{*}{\makecell[c]{DensePose for S\\ ResNet/Inception/NasNet for S\\ ResNet for F}} & \multirow{3}{*}{Average} & \multirow{3}{*}{Score} &  \multirow{3}{*}{\makecell[c]{VGG Face2\\RAF-DB}} & \textit{val} & 0.517  \\ \cline{13-14}   
		& & & & & & & &  & & &&  \multirow{2}{*}{\textit{test}} & \multirow{2}{*}{\textcolor{red}{\textbf{0.416}}} \\ 
		&  & & & & & & &  & & &&  &\\
		
		\cline{3-14} 
		
		&  &\multirow{3}{*}{\cite{zhu2019automatic}} & \multirow{3}{*}{\checkmark}  & \multirow{3}{*}{\checkmark} &  \multirow{3}{*}{\checkmark} & \multirow{3}{*}{} & \multirow{3}{*}{} & \multirow{3}{*}{\makecell[c]{VGG+SVR for F\\ Efficient+SVR for P\\ Densenet+SVR for S}} & \multirow{3}{*}{\makecell[c]{Grid\\Search}} & \multirow{3}{*}{Score} &\multirow{3}{*}{\makecell[c]{FER2013\\Emotic}} & \textit{val} &  0.672  \\ \cline{13-14}   
		& & & & & & & &  & & &&  \multirow{2}{*}{\textit{test}} & \multirow{2}{*}{0.444}\\
		&  & & & & & & &  & & &&  &
		\\ \hline
		
		% VGAF-1
		\multirow{10}{*}{\makecell[c]{VGAF\\(2020)}} & \multirow{10}{*}{3} & \multirow{2}{*}{\cite{petrova2020group}} & \multirow{2}{*}{}  & \multirow{2}{*}{\checkmark} &  \multirow{2}{*}{} & \multirow{2}{*}{} & \multirow{2}{*}{} & \multirow{2}{*}{VGG+ML} & \multirow{2}{*}{-} & \multirow{2}{*}{-} & \multirow{2}{*}{ImageNet} & \textit{val} & 57.18\%\\ \cline{13-14}   
		& & & & & & & &  & & &&  \textit{test} & 59.13\% \\ \cline{3-14}
		
		&  &\multirow{2}{*}{\cite{ottl2020group}} & \multirow{2}{*}{}  & \multirow{2}{*}{} &  \multirow{2}{*}{\checkmark} & \multirow{2}{*}{} & \multirow{2}{*}{} & \multirow{2}{*}{\makecell[c]{DeepSpectrum+AlexNet\\+VGG+DenseNet}} & \multirow{2}{*}{Mean} & \multirow{2}{*}{Score} & \multirow{2}{*}{-} & \textit{val} &  58.09\%  \\ \cline{13-14}    
		& & & & & & & &  & & &&  \textit{test} & 62.70\% \\ \cline{3-14}
		
		% 包括object
		&  &\multirow{2}{*}{\cite{liu2020group}} & \multirow{2}{*}{\checkmark}  & \multirow{2}{*}{\checkmark} &  \multirow{2}{*}{\checkmark} & \multirow{2}{*}{\checkmark} & \multirow{2}{*}{\checkmark} & \multirow{2}{*}{\makecell[c]{TSM for T\\Dense for F, OpenSmile for A}} & \multirow{2}{*}{Average} & \multirow{2}{*}{Score} &  \multirow{2}{*}{FER2013} & \textit{val} &  74.28\%  \\ \cline{13-14}   
		& & & & & & & &  & & &&  \textit{test} & \textcolor{red}{\textbf{76.85\%}} \\ \cline{3-14} 
		
		&  & \multirow{2}{*}{\cite{wang2020implicit}} & \multirow{2}{*}{}  & \multirow{2}{*}{\checkmark} &  \multirow{2}{*}{} & \multirow{2}{*}{\checkmark} & \multirow{2}{*}{\checkmark} & \multirow{2}{*}{K-injection} & \multirow{2}{*}{Cont} & \multirow{2}{*}{Feature} & \multirow{2}{*}{-} & \textit{val} &  66.19\%  \\ \cline{13-14}   
		& & & & & & & &  & & &&  \textit{test} & 66.40\% \\ \cline{3-14} 
		
		&  & \multirow{2}{*}{\cite{sun2020multi}} & \multirow{2}{*}{\checkmark}  & \multirow{2}{*}{\checkmark} &  \multirow{2}{*}{\checkmark} & \multirow{2}{*}{} & \multirow{2}{*}{\checkmark} & \multirow{2}{*}{\makecell[c]{TSM\\TBN}} & \multirow{2}{*}{\makecell[c]{Weight\\sum}} & \multirow{2}{*}{Score} & \multirow{2}{*}{ImageNet} & \textit{val} &  71.93\%  \\ \cline{13-14}  
		& & & & & & & &  & & && \textit{test} & 70.77\% \\ \hline
		
		% VGAF-2023
		\multirow{4}{*}{\makecell[c]{VGAF\\(2023)}}  & \multirow{4}{*}{3} & \multirow{2}{*}{\cite{li2023audio}} &  \multirow{2}{*}{\checkmark}  & \multirow{2}{*}{\checkmark} &  \multirow{2}{*}{} & \multirow{2}{*}{\checkmark} & \multirow{2}{*}{\checkmark} & \multirow{2}{*}{\makecell[c]{ResNet for F, SeNet for S\\Hubert large for A}} & \multirow{2}{*}{Cont} & \multirow{2}{*}{Feature}& \multirow{2}{*}{\makecell[c]{FER2013\\ImageNet}} & \textit{val} & 68.41\%\\ \cline{13-14}   
		& & & & & & & &  & & &&  \textit{test} & 72\% \\  \cline{3-14}
		
		&  & \multirow{2}{*}{\cite{augusma2023multimodal}} & \multirow{2}{*}{}  & \multirow{2}{*}{\checkmark} &  \multirow{2}{*}{} & \multirow{2}{*}{\checkmark} & \multirow{2}{*}{\checkmark} & \multirow{2}{*}{\makecell[c]{ViT-large for V\\ CNN+Transformer for A}} & \multirow{2}{*}{Average} & \multirow{2}{*}{Feature} & \multirow{2}{*}{ImageNet} & \textit{val} &  78.72\%  \\ \cline{13-14}   
		& & & & & & & &  & & &&  \textit{test} & \textcolor{red}{\textbf{75.13\%}}  \\ \hline
		
		\midrule
		\multicolumn{14}{p{32em}}{Prot.: Protocol; Cate.: Category.} \\
		\multicolumn{14}{p{25.14em}}{TSM: Temporal Shift Module; TBN: Temporal Binding Network; CAN: Cascade Attention Network.} \\
		\multicolumn{14}{p{32em}}{F: Face, S: Scene, P: Pose/skeleton, A: Audio, T: Temporal.} \\
		\multicolumn{14}{p{32em}}{Concat: Concatenation.}\\
		\multicolumn{14}{p{60em}}{* For HAPPEI and Group Coheison, RMSE is used as performance metric, while for other databases, accuracy is used.}
	\end{tabular}
\end{table*}

\begin{table*}
\centering
\caption{Performance comparison of remarkable deep learning techniques in image-based group-level emotion databases.}
\label{tab:method}
\scriptsize
%\vspace{0.2cm}
\begin{tabular}{|c|c|cccc|c|c|c|c|c|c|}
	\hline
	\multirow{2}{*}{Method} & \multirow{2}{*}{Database} & \multicolumn{4}{c|}{Modality} & \multirow{2}{*}{Network} & \multirow{2}{*}{\makecell[c]{Fusion\\ Scheme}} & \multirow{2}{*}{\makecell[c]{Fusion \\Stage}} & \multirow{2}{*}{Prot.} & \multirow{2}{*}{Cate.} & \multirow{2}{*}{Perf.} \\ \cline{3-6}
	&   & F & S & P & O &    &   &   &  &  &    \\
	
	%% GAF2.0
	\hline
	\multirow{2}{*}{\cite{shamsi2018group}} & \multirow{8}{*}{GAF2.0} & \multirow{2}{*}{\checkmark} & \multirow{2}{*}{} & \multirow{2}{*}{} & \multirow{2}{*}{} & \multirow{2}{*}{AlexNet} & \multirow{2}{*}{HeatMap} & \multirow{2}{*}{Feature} & \multirow{2}{*}{\textit{val}} & \multirow{2}{*}{3} & \multirow{2}{*}{55.23\%}\\ 
	
	& & & & & & & & & & & \\  
	\cline{1-1} \cline{3-12}
	
	\multirow{2}{*}{\cite{liu2018enhancing}} & \multirow{2}{*}{} & \multirow{2}{*}{\checkmark}  &  \multirow{2}{*}{\checkmark} & \multirow{2}{*}{}  & \multirow{2}{*}{} & \multirow{2}{*}{\makecell[c]{Xception for face \\ VGG for scene}} & \multirow{2}{*}{Concat} & \multirow{2}{*}{Feature} & \multirow{2}{*}{\textit{val}} & \multirow{2}{*}{3} & \multirow{2}{*}{71.83\%\multirow{2}{*}}\\ 
	& & & & & & & & & & & \\  
	\cline{1-1} \cline{3-12}
	
	\multirow{2}{*}{\cite{huang2022group}} & \multirow{2}{*}{} & \multirow{2}{*}{\checkmark} & \multirow{2}{*}{} & \multirow{2}{*}{} & \multirow{2}{*}{} & \multirow{2}{*}{CNN, RVLBP} & \multirow{2}{*}{DMKL}  & \multirow{2}{*}{Kernel} & \multirow{2}{*}{\textit{val}} & \multirow{2}{*}{3} & \multirow{2}{*}{79.49\%}\\ 
	& & & & & & & & & & & \\  
	\cline{1-1} \cline{3-12}
	
	\multirow{2}{*}{\cite{yu2019group}} & & \multirow{2}{*}{\checkmark} & \multirow{2}{*}{\checkmark}  &  & & \multirow{2}{*}{\makecell[c]{MobileNet for scene \\ LSTM for face}} & \multirow{2}{*}{Average}  & \multirow{2}{*}{Score} & \multirow{2}{*}{1-fold} & \multirow{2}{*}{3} & \multirow{2}{*}{78\%}\\ 
	
	& & & & & & & & & & & \\  
	\hline
	
	%% GAF3.0
	\multirow{2}{*}{\cite{nagarajan2019group}} & \multirow{5}{*}{GAF3.0} & \multirow{2}{*}{\checkmark} &  &  & & \multirow{2}{*}{\makecell[c]{Inception/VGG for scene\\ VGG for face}} & \multirow{2}{*}{SVM} & \multirow{2}{*}{Score} &  \multirow{2}{*}{\textit{val}} & \multirow{2}{*}{3} & \multirow{2}{*}{70.1\%}\\ 
	& & & & & & & & & & & \\ 
	\cline{1-1} \cline{3-12}
	
	\multirow{3}{*}{\cite{li2020group}} & & \multirow{3}{*}{\checkmark} & \multirow{3}{*}{\checkmark} & \multirow{3}{*}{\checkmark} & & \multirow{3}{*}{\makecell[c]{VGG+LSTM for face\\Dense for Skeleton\\Attention for Scene}}  & \multirow{3}{*}{Concat} & \multirow{3}{*}{Feature} & \multirow{3}{*}{\textit{val}} & \multirow{3}{*}{3} & \multirow{3}{*}{62.90\%}\\ 
	& & & & & & & & & & & \\ 
	& & & & & & & & & & & \\ 
	\hline

	\multirow{3}{*}{\cite{huang2019analyzing}} & \multirow{3}{*}{\makecell[c]{MultiEmoVA \\ HAPPEI* \\ GAF2.0}} & \multirow{3}{*}{\checkmark} & & & & \multirow{3}{*}{RVLBP, VGG} & \multirow{3}{*}{SVM-CGAK}  & \multirow{3}{*}{Kernel} & \multirow{3}{*}{\makecell[c]{5-fold\\ 4-fold \\ \textit{val}}} & \multirow{3}{*}{\makecell[c]{5 \\ 6 \\ 3}} & \multirow{3}{*}{\makecell[c]{54.40\% \\ 0.4920 \\ 72.17\%}}\\
	& & & & & & & & & & & \\ 
	& & & & & & & & & & & \\ 
	\hline
	
	\multirow{3}{*}{\cite{fujii2020hierarchical}} & \multirow{3}{*}{\makecell[c]{GAF2.0\\GAF3.0}} & \multirow{3}{*}{\checkmark}  & \multirow{3}{*}{\checkmark} & & \multirow{3}{*}{\checkmark} &  \multirow{3}{*}{\makecell[c]{VGG, attention for face\\VGG, attention for object\\VGG for scene}}&  \multirow{3}{*}{Hierarchical} & \multirow{3}{*}{Feature} & \multirow{3}{*}{\textit{val}} & \multirow{3}{*}{3} & \multirow{3}{*}{\makecell[c]{80.41\%\\ 76.61\%}}\\
	& & & & & & & & & & & \\ 
	& & & & & & & & & & & \\

	\hline
	\multirow{3}{*}{\cite{zhu2023towards}} & \multirow{3}{*}{\makecell[c]{MultiEmoVA \\ GAF2.0\\ GAF3.0}} &  \multirow{3}{*}{\checkmark} & \multirow{3}{*}{\checkmark} &  & \multirow{3}{*}{\checkmark} & \multirow{3}{*}{\makecell[c]{ResNet for face \\ VGG for object \\ VGG for scene}} & \multirow{3}{*}{UAL} & \multirow{3}{*}{Loss} & \multirow{3}{*}{\makecell[c]{5-fold\\ \textit{val}\\ \textit{val}}} & \multirow{3}{*}{\makecell[c]{5\\3\\3}} & \multirow{3}{*}{\makecell[c]{61.22\%\\79.19\%\\ 77.10\%}} \\ % F1
	& & & & & & & & & & & \\ 
	& & & & & & & & & & & \\ 
	
	\hline
	\multirow{4}{*}{\cite{guo2020graph}} & \multirow{4}{*}{\makecell[c]{GroupEmoW\\ GAF2.0\\SocEID}} & \multirow{4}{*}{\checkmark} & \multirow{4}{*}{\checkmark} & \multirow{4}{*}{\checkmark} & \multirow{4}{*}{\checkmark} & \multirow{4}{*}{\makecell[c]{VGG for face \\ SE-ResNet for skeleton \\ SENet for object \\ Inception for scene }} & \multirow{4}{*}{GNN} & \multirow{4}{*}{Score}  & \multirow{4}{*}{\makecell[c]{\textit{test}\\ \textit{val}\\\textit{test}}} & \multirow{4}{*}{\makecell[c]{3\\3\\8}} & \multirow{4}{*}{\makecell[c]{89.14\%\\ 78.16\%\\ 91.61\%}}\\
	& & & & & & & & & & & \\ 
	& & & & & & & & & & & \\ 
	& & & & & & & & & & & \\ 
	
	\hline
	\multirow{2}{*}{\cite{khan2021regional}} & \multirow{2}{*}{\makecell[c]{GAF2.0 \\ GroupEmoW}} & \multirow{2}{*}{\checkmark} & \multirow{2}{*}{\checkmark} &\multirow{2}{*}{} & \multirow{2}{*}{\checkmark} & \multirow{2}{*}{\makecell[c]{Resnet for all modalities\\ Attention module}} & \multirow{2}{*}{\makecell[c]{CARAN}} & \multirow{2}{*}{Loss} & \multirow{2}{*}{\makecell[c]{\textit{val}\\ \textit{test}}} & \multirow{2}{*}{3} & \multirow{2}{*}{\makecell[c]{67.61\%\\ 90.18\%}}\\
	& & & & & & & & & & & \\ 
	\hline
	\multirow{3}{*}{\cite{zhang2022semi}} & \multirow{3}{*}{\makecell[c]{GroupEmoW\\ GAF2.0\\ GAF3.0}} & \multirow{3}{*}{\checkmark} & \multirow{3}{*}{\checkmark} & &  & \multirow{3}{*}{ResNet for all modalities} & \multirow{3}{*}{FusionNet} & \multirow{3}{*}{Loss} & \multirow{3}{*}{\makecell[c]{\textit{test}\\\textit{val}\\ \textit{val}}} & \multirow{3}{*}{3} & \multirow{3}{*}{\makecell[c]{88.67\%\\ 78.51\%\\ 77.01\%}}\\
	& & & & & & & & & & & \\ 
	& & & & & & & & & & & \\ 
	
	\hline
	\multirow{3}{*}{\cite{wang2022congnn}} & \multirow{3}{*}{\makecell[c]{GroupEmoW\\ GAF2.0\\ GAF3.0}} & \multirow{3}{*}{\checkmark} & \multirow{3}{*}{\checkmark} & & \multirow{3}{*}{\checkmark} & \multirow{3}{*}{\makecell[c]{ResNet+LSTM+GNN for face \\ SE-ResNet+GNN for object \\ SE-ResNet+GNN for scene}} & \multirow{3}{*}{ECL} & \multirow{3}{*}{Loss} & \multirow{3}{*}{\makecell[c]{\textit{test}\\\textit{val}\\\textit{val}}} & \multirow{3}{*}{3} & \multirow{3}{*}{\makecell[c]{90.06\%\\ 79.45\%\\ 79.95\%}}\\
	& & & & & & & & & & & \\ 
	& & & & & & & & & & & \\ 
	\hline
	\multirow{2}{*}{\cite{xie2023most}} & \multirow{2}{*}{\makecell[c]{GAF3.0\\GroupEmoW}} & \multirow{2}{*}{\checkmark} & \multirow{2}{*}{\checkmark} &  & & \multirow{2}{*}{Multi-scale Transformer} & \multirow{2}{*}{DCAT} & \multirow{2}{*}{Feature} &  \multirow{2}{*}{\makecell[c]{\textit{val} \\ \textit{test}}} & \multirow{2}{*}{3} & \multirow{2}{*}{\makecell[c]{79.20\%\\ 90.47\%}}\\
	& & & & & & & & & & & \\ 
	\hline
	
	\multirow{2}{*}{\cite{quach2022non}} & \multirow{2}{*}{\makecell[c]{GECV-GroupImg\\ GAF3.0}} & \multirow{2}{*}{\checkmark} & & & & \multirow{2}{*}{EmoNet} & \multirow{2}{*}{NVPF} & \multirow{2}{*}{Feature} & \multirow{2}{*}{\textit{val}} & \multirow{2}{*}{3} & \multirow{2}{*}{\makecell[c]{77.02\%\\ 76.12\%}}\\
	& & & & & & & & & & & \\ 
	\hline
	\midrule
	\multicolumn{12}{p{32em}}{Prot.: Protocol; Cate.: Category.} \\
	\multicolumn{12}{p{70em}}{NVPF: Non-volume Preserving-based Fusion; DMKL: Deep Multiple Kernel Learning; GNN: Graph Neural Network.} \\
	\multicolumn{12}{p{70em}}{UAL: Uncertain-aware Learning; CARAN: Context-aware Regional Attention Network; Concat: Concatenation.}\\
	\multicolumn{12}{p{70em}}{ECL: Emotion context-consistent learning; DCAT: Dual-branch Cross-Patch Attention Transformer.}\\
	\multicolumn{12}{p{32em}}{F: Face, S: Scene, P: Pose/skeleton, O: Object.} \\
	\multicolumn{12}{p{60em}}{* For HAPPEI, RMSE is used as performance metric, while for other databases, accuracy is used.}
\end{tabular}
\end{table*}

\begin{table*}
\centering
\caption{Performance comparison of remarkable deep learning techniques in video-based group-level emotion (VGAF) database, where F, S, P, and A mean face, scene, pose, and audio modality, respectively.}
\label{tab:VGAF}
\scriptsize
%\vspace{0.2cm}
\begin{tabular}{|c|c|cccc|c|c|c|c|c|c|}
	\hline
	\multirow{2}{*}{Method} & \multirow{2}{*}{Dataset} & \multicolumn{4}{c|}{Modality} & \multirow{2}{*}{Network} & \multirow{2}{*}{\makecell[c]{Fusion \\Scheme}} & \multirow{2}{*}{\makecell[c]{Fusion \\Stage}} & \multirow{2}{*}{Prot.} & \multirow{2}{*}{Cate.} & \multirow{2}{*}{Accuracy}\\ \cline{3-6}
	& & F & S & P & A &  &  &  &  &  & \\
	\hline
	
	\multirow{2}{*}{Sharma+\cite{sharma2019automatic}} & \multirow{6}{*}{VGAF} & \multirow{2}{*}{} & \multirow{2}{*}{\checkmark} & \multirow{2}{*}{} & \multirow{2}{*}{\checkmark} & \multirow{2}{*}{\makecell[c]{LSTM for scene\\OpenSMILE for audio}}  & \multirow{2}{*}{Concat} & \multirow{2}{*}{Feature} & \multirow{2}{*}{\textit{val}} & \multirow{2}{*}{3} & \multirow{2}{*}{47.50\%}\\ 
	
	& & & & & & & & &  &  &  \\
	\cline{1-1} \cline{3-12}
	%Liu+~\cite{liu2020group} &  &  &  &  & \checkmark & LSTM, OpenSMILE  & Concat & Feature  & \textit{test} & 3 & 76.85\%\\
	%\cline{1-1} \cline{3-12}
	%Wang+\cite{wang2020implicit} &  & F & SC & SK & Obj &  & &  & \textit{test} & 3 & 66.40\%\\
	%\cline{1-1} \cline{3-12}
	\multirow{2}{*}{Pinto+\cite{pinto2020audiovisual}} & \multirow{2}{*}{} & \multirow{2}{*}{} & \multirow{2}{*}{\checkmark} & \multirow{2}{*}{} & \multirow{2}{*}{\checkmark} & \multirow{2}{*}{\makecell[c]{ResNet for scene \\ Bi-LSTM for audio}} & \multirow{2}{*}{SVM} & \multirow{2}{*}{Score} & \multirow{2}{*}{\textit{val}} & \multirow{2}{*}{3} & \multirow{2}{*}{65.74\%}\\
	
	& & & & & & & & &  &  &  \\
	\cline{1-1} \cline{3-12}
	%Sun+\cite{sun2020multi} &  & F & SC & SK & Obj &  & &  & \textit{test} & 3 & 70.77\%\\
	%\cline{1-1} \cline{3-12}
	%Ottl+\cite{ottl2020group} &  & F & SC & SK & Obj &  & &  & \textit{test} & 3 & 62.70\%\\
	%\cline{1-1} \cline{3-12}
	%Petrova+\cite{petrova2020group} &  & F & SC & SK & Obj &  & &  & \textit{test} & 3 & 59.13\%\\
	%\cline{1-1} \cline{3-12}
	%Augusma+\cite{augusma2023multimodal} &  & F & SC & SK & Obj &  & &  & \textit{test} & 3 & 75.13\%\\
	%\cline{1-1} \cline{3-12}
	\multirow{2}{*}{Evtodienko+\cite{evtodienko2021multimodal}} & \multirow{2}{*}{} & \multirow{2}{*}{} & \multirow{2}{*}{\checkmark} & \multirow{2}{*}{} & \multirow{2}{*}{\checkmark} & \multirow{2}{*}{\makecell[c]{Hubert+Attention for audio\\ResNet+Attention for scene}} & \multirow{2}{*}{Concat} & \multirow{2}{*}{Feature} & \multirow{2}{*}{\textit{val}} & \multirow{2}{*}{3} & \multirow{2}{*}{60.37\%}\\
	& & & & & & & & &  &  &  \\
	\cline{1-1} \cline{3-12}
	% \multirow{2}{*}{Slogrove+\cite{slogrove2022group}} & \multirow{2}{*}{} & \multirow{2}{*}{} & \multirow{2}{*}{} & \multirow{2}{*}{\checkmark} &  & \multirow{2}{*}{LSTM} & \multirow{2}{*}{-} & \multirow{2}{*}{-} & \multirow{2}{*}{\textit{val}} & \multirow{2}{*}{3} & \multirow{2}{*}{91\%}\\
	% & & & & & & & & &  &  &  \\
	\hline
	\multirow{2}{*}{Quach+\cite{quach2022non}} & \multirow{2}{*}{\makecell[c]{GECV\\-GroupVid}} & \multirow{2}{*}{\checkmark} & \multirow{2}{*}{} &  \multirow{2}{*}{} & \multirow{2}{*}{} &  \multirow{2}{*}{EmoNet} & \multirow{2}{*}{TNVPF} & \multirow{2}{*}{Feature} & \multirow{2}{*}{\textit{test}} & \multirow{2}{*}{3} & \multirow{2}{*}{70.97\%}\\
	& & & & & & & & &  &  &  \\
	\hline
	\midrule
	\multicolumn{12}{p{32em}}{Prot.: Protocol; Cate.: Category.} \\
	\multicolumn{12}{p{32em}}{TNVPF: Temporal Non-volume Preserving-based Fusion.} \\
	\multicolumn{12}{p{32em}}{Concat: Concatenation.}\\
	\multicolumn{12}{p{32em}}{F: Face, S: Scene, P: Pose/skeleton, A: Audio.} \\
\end{tabular}
\end{table*}

\section{Challenges and future direction}

%\textcolor{red}{group-level emotion recognition in the wild has received much attention in computer vison community. It is a very challenge issue, due to interactions taking place between various numbers of people, different occlusion.}

Group-level emotion recognition in unconstrained environments has become significant attention within the computer vision community, offering substantial implications for social public security and education. This article delves into the intricate concepts of group dynamics and emotion, along with methodologies for recognizing group-level emotion, and pertient datasets, aiming to provide a comprehensive analysis of the current landscape and future trajectories of GER. This endeavor furnishes a robust theoretical foundation for potential applications of GER in domains such as social psychology, human-computer interaction, and smart cities. In this section, we summarize and discuss the future prospects of GER from three pivotal dimensions: database-level, technique-level, multimodality, and evaluation metric.

GER encounters three primary technical challenges. Firstly, the accurate discernment of  emotions inherently presents complex, compounded by the potential bias introduced by human subjective labeling in annotating datasets. Secondly, the fluctuating number of individuals and diverse scenes within a group necessitates enhanced generalization and robustness of features. Furthermore, different extracted features may convey inconsistent emotions. Lastly, the cross-fusion of diverse features in a multimodal model and achieving end-to-end feature learning pose significant challenges. Despite these obstacles hindering GER advancement, its potential applications span a wide spectrum.
%\textcolor{red}{However, GER presents three key technical challenges. accurately understanding emotions itself is a complex problem, and subjective labeling by humans may lead to bias in the annotated dataset; (2) the changes in the number of people and scenes in a group require higher requirements for generalization and robustness of features. On the other hand, the extracted different features may express inconsistent emotions; (3) the cross-fusion of different features in a multimodal model and how to achieve end-to-end feature learning are also challenging. Despite these technical challenges limiting the development of group-level emotion recognition, it still has a wide range of applications. }

\subsection{GER Database}
While currently available group emotion databases primarily collect images and videos from websites and media platforms, their sample size remain relatively small. On the other hand, Smith~\etal~\cite{smith2015dynamics} emphasize the significance of the changes in group emotions along time, noting that individuals may react differently to external group members based on their prevailing emotion states. Pantic~\etal~\cite{pantic2006dynamics} found that dynamic videos provide more discriminative information to extract temporal changes in emotions. Therefore, the collection of richer and more realistic video data holds promise for furnishing contextual semantic insights into group interactions and dynamic processes, facilitating a more nuanced observation of emotional dynamics within team.

Presently, database annotations are typically derived from independent observers rating the images, lacking subjective evaluation like self-reporting measurements. This limitation may lead to the challenges in acquiring such measurements from online media platforms. However, the absence of subjective evaluation may potentially yield considerable performance in group-level emotion recognition. By integrating subjective evaluation and other auxiliary information, computer scientists can devise advanced methodologies for analyzing data, bridging the gap between computational methods and social science research. Moreover, existing databases primarily reply on methods such as multi-observer cross-calibration to categorize images or videos, which may introduce calibration biases due to cultural disparities and overlook the fundamental tenets of group emotion posited by social psychologists~\cite{barsade2007does}. Therefore, it is crucial to expanding the repertoire of basic emotion categories to encompass more generalized emotion states in affective computing is imperative. Additionally, involving social psychologists in the data collection and annotation process can furnish more rational and valuable calibration information.

\subsection{GER Technique}

The methods ranging from Convolutional Neural Networks (CNN) to Recurrent Neural Networks (RNN), Hybrid Networks combining CNN and RNN, and Graph Convolutional Networks (GCN), have already demonstrated remarkable success in GER. However, the challenge of limited sample sizes necessitates the exploration of unsupervised and self-supervised learning techniques. Unsupervised learning technique like Generative adversarial networks~\cite{xia2021local} and generative AI~\cite{latif2023generative} can be valuable for learning meaningful representations from unannotated data, enabling the development of robust deep GER models. These techniques help in extracting rich features from data, even when labeled examples are scarce. Similarly, self-supervised learning paradigms, such as contrastive learning~\cite{kim2022emotion}, offer a way to learn representations from auxiliary tasks, enhancing the generalization capabilities of GER models across diverse group contexts. %By harnessing the power of unsupervised and self-supervised learning, future deep learning models will be capable of extracting rich and transferable representations of group-level emotions, even in data-scarce scenarios.

As GER systems move towards real-world deployment, the demand for continual learning and adaptive capabilities become increasingly critical. Deep learning architectures will need to evolve to accommodate dynamic changes in group compositions, social contexts, and environmental conditions over time. Incremental learning strategies, lifelong learning approaches, and adaptive neural networks will play crucial roles in enabling models to adapt and refine their representations based on incoming data streams. Moreover, techniques for mitigating catastrophic forgetting and domain adaptation will be essential for ensuring the long-term stability and effectiveness of GER systems. These methods help models retain previously learned knowledge while adapting to new information, thereby enhancing their robustness in real-world scenarios.

By embracing unsupervised and self-supervised learning techniques, as well as continual learning and adaptive methodologies, future deep learning systems will be better equipped to capture the nuanced dynamics of group-level emotions across diverse scenarios and domains. This holistic approach holds the potential to significantly advance the field of Group Emotion Recognition and its applications in various domains.

\subsection{Multi-modal architecture for GER}
In the research process of GER, we have observed a continuous evolution towards richer and more diverse  in feature representations. Initially, the focus was predominantly on a single facial feature. However, as research progressed, there was a shift towards utilizing multiple features concurrently, including facial features, local object features, and scene features. This enrichment of feature diversity has contributed to an enhanced accuracy of group-level emotion recognition to a certain extent. Moreover, researcher have begun to explore the integration of different types of data sources. For example, studies by Sharma~\etal~\cite{sharma2019automatic}, Wang~\etal~\cite{wang2020implicit}, Liu~\etal~\cite{liu2021multimodal}, and Pinto~\etal~\cite{pinto2020audiovisual} have incorporated  both video frame features and audio features, while Liu~\etal~\cite{liu2020group} and Sun~\etal~\cite{sun2020multi} have combined both facial and audio features. These approaches, known as multimodal methods, offer advantages such as robustness against interference, high interpretability, and broad applicability compared to single-modal methods. 

Traditionally, GER relied primarily on single-modal information. However, as the field has progressed and dataset forms have diversified, researchers have increasingly delved into multi-modal methods. These methods leverage various data forms including images, videos, and sounds. The adoption of multimodal fusion techniques in group-level emotion recognition not only facilitates a more comprehensive understanding of emotions but also enhances the accuracy and robustness of the recognition process.

Nevertheless, the application of multi-modal fusion in GER poses several challenges. Obtaining and annotating datasets for group-level emotion recognition, particularly those encompassing different modalities, can be arduous. Furthermore, variations in feature extraction and fusion methods across modalities present additional complexities, making it challenging to harmonize features between different modalities.

Overall, while multi-modal fusion encounters challenges in GER, its potential and advantages are undeniable. With advancements in technology and ongoing research efforts, it is anticipated that these challenges will be gradually addressed, and multimodal methods will emerge as the primary research direction in the field of GER. Future research endeavors will likely focus on effectively integrating information from diverse modalities and constructing larger and more diverse datasets for GER, thereby presenting both challenges and opportunities in this dynamic field.

%, we can see from the main stages of ~\textit{et al.} we discussed above that more and more researchers are deeply exploring multimodal methods. Images, videos, sounds, and other data forms are all used in multimodal methods. The application of multimodal fusion methods in the field of ~\textit{et al.} can not only provide more comprehensive emotional understanding, but also improve the accuracy and robustness of ~\textit{et al.}. At the same time, there are also some challenges in applying multimodal fusion to ~\textit{et al.}. For example, multimodal fusion requires different forms of datasets, and obtaining and annotating ~\textit{et al.} datasets is relatively difficult. There are differences in feature extraction and fusion methods for different modalities, and it is difficult to adapt the features between modalities.

%Overall, although multimodal fusion faces some challenges in ~\textit{et al.}, its potential and advantages cannot be ignored. With the advancement of technology and the deepening of research, we have reason to believe that these challenges will be gradually overcome, and multimodal methods will become the main research direction in the field of ~\textit{et al.}. Future research will focus more on how to effectively integrate information from different modalities, as well as how to construct larger and richer datasets for ~\textit{et al.}. This is a field full of challenges and opportunities.

\subsection{Evaluation metric of GER}

It is worth noting that while accuracy is a common metric in GER, it can be influenced by biased data. To address this issue, F1-score offers a more comprehensive evaluation by considering True Positives (TP), False Positives (FP), and False Negatives (FN). This metric provides a balanced assessment of the true classification performance.

%It is important to note that while accuracy is commonly used in GER. it is susceptible to bias data. F1-score addresses this bias issue by considering the total True Positives (TP), False Positives (FP) and False Negatives (FN), providing a more comprehensive measure of the true classification performance.

Although the data may not exhibit severe imbalance, as indicated in Table~\ref{tab:databaseintro}, it may be more suitable to employ metrics such as Unweighted F1-score (UF1) and Unweighted Average Recall (UAR) to assess method performance. UF1, also known as macro-averaged F1-score, calculates the average F1-score across all classes, offering equal weighting to each class in multi-class scenarios. Conversely, UAR computes the average accuracy per class, normalized by the total number of classes. UAR helps mitigate bias arising from class imbalance existing in some databases,\eg, SiteGroEmo and is often referred to as balanced accuracy.

%Although the data imbalance is not severe, as shown in Table~\ref{tab:databaseintro}, it may be more appropriate to utilize Unweighted F1- score (UF1) and Unweighted Average Recall (UAR) to measure the performance of various methods. UF1, also known as macro-averaged F1-score, is determined by averaging the per-class F1-scores, providing equal emphasis on rare classes in imbalanced multi-class settings. On the other hand, UAR is defined as the average accuracy of each class divided by the number of classes, without consideration of samples per class. UAR can reduce the bias caused by class imbalance and is commonly referred to as balanced accuracy.

\ifCLASSOPTIONcaptionsoff
\newpage
\fi

\footnotesize
\bibliographystyle{IEEEtran}
\bibliography{sn-bib-ger}

\vfill

\end{document}